\documentclass[a4paper,11pt,reqno]{amsart}
\usepackage{amsmath,amssymb,amsthm} 
\usepackage{graphics} 
\usepackage{neuralnetwork}
\usetikzlibrary{shapes}

\usepackage[colorlinks = true]{hyperref}

\usepackage[hyperref,doi,url=false,isbn=false, giveninits=true, backref=true,style=alphabetic,maxbibnames=99, backend=biber]{biblatex}
 \bibliography{deep_learning.bib}

\numberwithin{equation}{section}

\newtheorem{theorem}{Theorem}[section]
\newtheorem{lemma}[theorem]{Lemma}

\theoremstyle{definition}

\newtheorem{remark}[theorem]{Remark}

\newcommand{\R}{\mathbb{R}}

\newcommand{\be}{\begin{equation}}
\newcommand{\ee}{\end{equation}}

\makeatletter
\renewcommand{\fnum@figure}{Fig. \thefigure}
\makeatother

\def\Id{{\rm Id}}
\def\tr{{\rm tr}}

\def\EE{\mathbb{E}}
\def \W{\mathcal{W}}
\def\PP{\mathbb{P}}

\newcommand{\bra}[1]{\left( #1 \right)}
\newcommand{\sqa}[1]{\left[ #1 \right]}
\newcommand{\cur}[1]{\left\{ #1 \right\}}
\newcommand{\ang}[1]{\left< #1 \right>}
\newcommand{\abs}[1]{\left| #1 \right|}
\newcommand{\nor}[1]{\left\| #1 \right\|}

\usepackage{enumerate}
\newcommand{\Lip}{\operatorname{Lip}}

\newcommand{\X}{\mathcal{X}}

\title[Quantitative Gaussian Approximation of Randomly Initialized DNN]{Quantitative Gaussian Approximation of Randomly Initialized Deep Neural Networks}

\author[A. Basteri]{Andrea Basteri}
\address{INRIA Paris, 75012, Paris, France. Département d’Informatique, École Normale Supérieure, Paris, France. PSL Research University, Paris, France}
\email{andrea.basteri@inria.fr}

\author[D. Trevisan]{Dario Trevisan}
\address{D.T.: Dipartimento di Matematica, Università degli Studi di Pisa, 56125 Pisa, Italy. Member of the INdAM GNAMPA group.}
\email{dario.trevisan@unipi.it}

\date{\today  }
\subjclass[2010]{60F05, 60G15, 68T99}
\keywords{Neural Networks, Central Limit Theorem, Wasserstein distance}

\begin{document}

\maketitle

\begin{abstract}
Given any deep fully connected  neural network, initialized with random Gaussian parameters, we bound from above the quadratic Wasserstein distance between its output distribution and a suitable Gaussian process. Our explicit inequalities indicate how the hidden and output layers sizes affect the Gaussian behaviour of the network and quantitatively recover the distributional convergence results in the wide limit, i.e., if all the hidden layers sizes become large.
\end{abstract}

\section{Introduction}

Contemporary machine learning has seen a surge in applications of deep neural networks, achieving state-of-the-art performances in many contexts, from speech and visual recognition \cite{lecun_deep_2015}, to feature extraction and generation of samples \cite{goodfellow_generative_2014}. Such a deep learning revolution \cite{sejnowski_deep_2018} is often compared to the advent of the industrial age \cite[Chapter 0]{roberts2022principles}, to the extent that the search for powerful and efficient steam machines historically led to the first theoretical investigations  in thermodynamics. This analogy may be a bit of a stretch, but the fundamental questions about the scope and limits of deep learning methods have indeed stimulated new mathematical research in various fields, most notably probability, statistics and statistical physics, with exciting recent progresses towards enlightening answers.

\subsection{Random neural networks} A successful approach focuses on the scaling limit of large neural networks whose parameters are randomly sampled. Besides the introduction of several mathematical tools, there are convincing reasons for such a study. Firstly, a Bayesian approach to machine learning problems would require a prior distribution on the set of networks, which should then be updated via Bayes rule, given the observations, e.g., the training set in supervised learning. In practice, however, large neural networks are trained via iterative optimisation algorithms \cite{Goodfellow-et-al-2016}, which still require a careful initialization of the parameters, often suggested to be randomly sampled \cite{narkhede2022review}, thus providing a second motivation for such a study. Even more strikingly, training only the parameters in the last layer of a randomly initialized neural network may still provide good performances and of course an extreme speed-up in the training algorithm, a fact that stimulated a whole machine learning paradigm \cite{cao2018review}.

In view of the above reasons, the study of random neural networks is thus not new, and in fact dates back to pioneers in the field \cite{rosenblatt1958perceptron}. In the seminal work \cite{neal_priors_1996}, motivated by the Bayesian approach, Neal first proved that wide shallow networks, i.e., with a large number of parameters but  only one hidden layer, converge to Gaussian processes, if the parameters are random and suitably tuned, as a consequence of the Central Limit Theorem (CLT). More recently, in \cite{matthews_gaussian_2018, lee_deep_2018}, convergence has been extended to deeper networks, stemming a series of works that aim to explain the success story of over-parametrized neural networks, totally unexpected from traditional statistical viewpoints, by linking it to the well-established theory of Gaussian processes \cite{williams2006gaussian}. A breakthrough is the realization \cite{lee_wide_2019} that the training dynamics, e.g.\ via gradient descent, can be also traced in the scaling limit and become a differential equation of gradient flow type associated to the loss functional and the so-called Neural Tangent Kernel, i.e., a quadratic form associated to the gradient  with respect to the parameters of the network, evaluated on the training set.  Let us also mention the alternative perspective of the so-called mean field limit of networks \cite{mei_2018, nguyen_rigorous_2021}, where parameters are tuned in a Law of Large Numbers regime, leading to deterministic non-linear dynamics for the network training.

\subsection{Our contribution} In this work, we provide a  quantitative proof of the Gaussian behaviour of deep neural networks with random parameters at initialization, i.e., without addressing the correlations induced by the training procedure. We thus complement the seminal works \cite{matthews_gaussian_2018} and subsequent ones focusing on convergence for wide networks \cite{yang_wide_2019, bracale_large-width_2020, hanin_random_2021}, by providing for the first time explicit rates for convergence in distribution in the case of deep networks. The metric we adopt is the Wasserstein distance, which generalizes the so-called earth mover's distance, and provide a natural interpretation of the discrepancy between the two distributions in terms of the average cost between transporting one over the other. Indeed, it was historically introduced by Monge as the Optimal Transport problem, i.e., to efficiently move a given distribution of mass from a location to another \cite{villani2009optimal}. The theory of optimal transport has grown in the last decades, with applications beyond calculus of variation and probability, from geometry and partial differential equations \cite{ambrosio2014metric}, to statistics and machine learning \cite{peyre2019computational}. We use in fact only basic properties of the Wasserstein distance, that are all recalled in Section~\ref{sec:wasserstein}, and focus for simplicity of the exposition only on the quadratic case, denoted by $\W_2$, i.e., when the transport cost between two locations is the squared Euclidean distance.

\begin{figure}
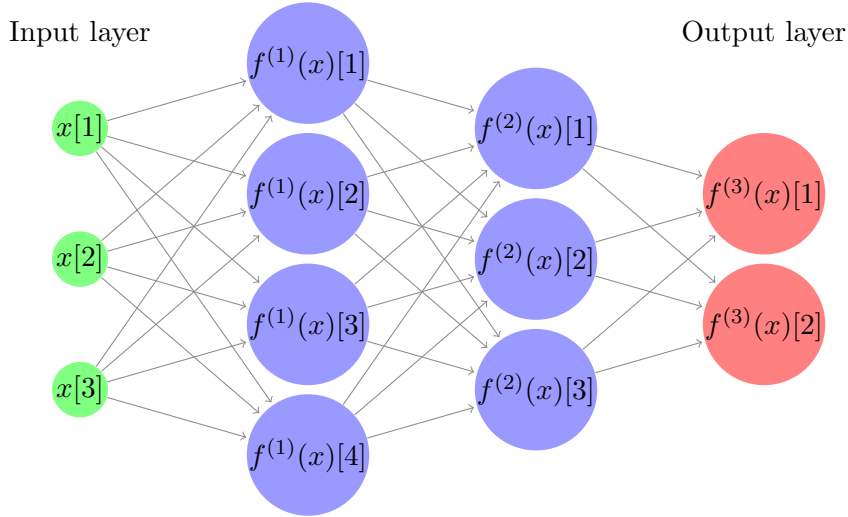

	\begin{neuralnetwork}[height=4, nodesize=17pt, nodespacing=4.5em, layerspacing=3cm, layertitleheight=1em]
		\newcommand{\nodetextclear}[2]{}
		\newcommand{\nodetextx}[2]{$x[#2]$}
		\newcommand{\nodetexthidden}[2]{$f^{(1)}(x)[#2]$}
				\newcommand{\nodetexthiddentwo}[2]{$f^{(2)}(x)[#2]$}
		\newcommand{\nodetexty}[2]{$f^{(3)}(x)[#2]$}
		\inputlayer[count=3, bias=false, title=Input layer, text=\nodetextx]
		\hiddenlayer[count=4, bias=false,  text=\nodetexthidden] \linklayers
		\hiddenlayer[count=3, bias=false,  text=\nodetexthiddentwo] \linklayers
		\outputlayer[count=2, title=Output layer, text=\nodetexty] \linklayers
	\end{neuralnetwork}
	\caption{Graphical representation \cite{cowan_battlesnakeneural_2022} of a fully connected neural network with $L=3$ layers, input size $n_0=3$, output size $n_3=2$ and hidden layer sizes $n_1=4$, $n_2=3$.}\label{fig:nn}
	\end{figure}

To further keep the exposition short and accessible, we discuss only so-called fully connected neural networks (or multi-layer perceptrons), where all units (neurons) in each layer are linked with those on the next layer, from the input layer to the output (see Figure~\ref{fig:nn}). We denote by $L+1$ the total number of layers (including input and output), with sizes $n_0$ for the input layer, $n_L$ for the output and $(n_1, \ldots, n_{L-1})$ for the intermediate hidden ones. Then, for every choice of the parameters, often called weights (matrices) ${\bf W} =(W^{(\ell)})_{\ell=0}^{L-1}$ and biases (vectors) ${\bf b} = (b^{(\ell)})_{\ell=0}^{L-1}$, and a (usually fixed) non-linear activation function $\sigma$ acting on each intermediate node (except for the last layer), a neural network $f^{(L)}: \R^{n_0} \to \R^{n_L}$ is defined (see Section~\ref{sec:nn} for a precise definition).

Using the concepts and notation mentioned above, our main result can be stated as follows (see Theorem~\ref{thm:main-induction} for a more precise version).

\begin{theorem}\label{thm:main-clean}
Consider a fully connected neural network $f^{(L)}: \R^{n_0} \to \R^{n_L}$ with Lipschitz activation function $\sigma$, and random
 weights ${\bf W}$ and biases ${\bf b}$ that are independent Gaussian random variables, centered with
\begin{equation}\label{eq:variances-clean} \EE\sqa{ (W^{(\ell)}_{i,j})^2} = \frac 1 {n_\ell}, \quad \EE\sqa{ (b^{(\ell)}_{i})^2} = 1, \quad \text{for every $\ell = 0,1, \ldots, L-1$ and all $i$, $j$.}\end{equation}
Then, for every given set  of $k$ inputs $\X = \cur{x_i}_{i=1}^k \subseteq \R^{n_0}$, the distribution of $f^{(L)}[\X] = (f^{(L)}(x_i))_{i=1}^k$, i.e., the law of the output of the random network evaluated at the inputs $\X$, is quantitatively close to a centered multivariate Gaussian  $\mathcal{N}(K^{(L)}[\X])$ with a (computable) covariance matrix $K^{(L)}[\X]$, in terms of the quadratic Wasserstein distance:
\begin{equation}\label{eq:thesis-clean} \W_2\bra{ \PP_{f^{(L)}[\X]},  \mathcal{N}(K^{(L)}[\X])} \le C \sqrt{n_L} \sum_{\ell=1}^{L-1} \frac{1}{\sqrt{n_\ell}},\end{equation}
where $C\in (0, \infty)$ is a constant depending on the activation function $\sigma$, the input $\X$ and the number of layers $L$ only, not on the hidden or output layer sizes $(n_\ell)_{\ell=1}^{L}$. 
\end{theorem}

The random network considered in our result is also known as the Xavier initialization, which sets the initial weights using a scaling factor that takes into account the number of input and output neurons in each layer. Let us point out that several other random distributions can be considered  for specific activation functions or network architectures, see \cite{narkhede2022review} for a review.

The matrix $K^{(L)}[\X]$ depends on the activation function $\sigma$, the input $\X$ and the output dimension $n_L$ (not on the hidden layer sizes $(n_\ell)_{\ell=1}^{L-1}$). It is indeed recursively computable (see Section \ref{sec:nngp}) and in some cases -- e.g.\ for ReLU activation --  closed formulas can be obtained, see \cite{cho2009kernel, lee_deep_2018}. An important feature is that all $n_L$ output neurons in the Gaussian approximation are independent and identically distributed variables (for any input) -- the desirability of such property is often discussed, but let us recall that we are considering here non-trained networks.

A useful feature of our result is that it explicitly indicates how the layer sizes affect the Gaussian behaviour of the output of the network. In particular, in the wide limit where all the hidden layers sizes $n_i$'s are large (keeping $n_L$ and $n_0$ fixed) the Wasserstein distance becomes infinitesimal and convergence in distribution for fixed input set follows. Some considerations apply as well in the narrow deep limit, since the constant $C$ in \eqref{eq:thesis-clean} can be made explicit and also more general variances in \eqref{eq:variances-clean} can be introduced: the contribution $\sqrt{n_L}/\sqrt{n_\ell}$ naturally associated to the $\ell$-th hidden layer is weighted by exponentially contracting (or growing) terms as $L$ becomes large (see Remark~\ref{rem:deep-limit}).

\subsection{Related literature}

Our proof is based on the natural idea, already present in \cite{neal_priors_1996}, that the Gaussian limit emerges due to a combination, in each layer, of the CLT scaling for the weights parameters and the inherited convergence in the previous layer, which almost yields independence for the neurons. For shallow networks, which correspond to the base case of our argument, exact independence holds and Gaussian behaviour is immediate from the CLT. When considering deeper architectures, the generalization is not immediate and \cite{matthews_gaussian_2018} made use of a CLT for triangular arrays. The arguments were later simplified, e.g.\ via the use of characteristic functions \cite{bracale_large-width_2020} or moments \cite{hanin_random_2021}, but quantitative convergence in terms of any natural distance (not only the Wasserstein one) was missing so far. It turns out in fact that using the Wasserstein distance allows for quantitatively tracking the hidden layers and yields a relatively straightforward proof by induction, where the triangle inequality is employed to obtain exact independence from the previous layers, and most terms are  estimated without much effort. 

Let us point out that our main result provides convergence for any finite set of $k$-inputs of a random deep neural network. Obtaining rates for deep networks in suitable functional spaces (informally, for infinitely many inputs) seems considerably more difficult, although useful, e.g.\ to study architectures with bottlenecks \cite{agrawal_wide_2020}, i.e., when some hidden layers that are purposely kept with small width, such as in autoencoders. The case of shallow networks has been addressed in \cite{eldan2021non, klukowski2022rate}, using Stein's method for the quadratic Wasserstein distance in the space of squared integrable functions on the sphere. In Section~\ref{sec:functional} we hint at how one could further extend our argument to obtain bounds in functional spaces.

\subsection{Extensions and future work}

As already mentioned, we choose to keep technicalities at minimum by considering only the quadratic Wasserstein distance, fully connected architectures, Gaussian weights and biases and  a finite subset of inputs. In view of the existing literature on the subject, we expect however that all such aspects may be relaxed, e.g., from Wasserstein distances of any order $p$ or further variants, such as the entropic regularized optimal transport \cite{peyre2019computational}; to more general network architectures, such as those of convolutional type \cite{borovykh_gaussian_2019, garriga-alonso_deep_2018}, following e.g.\ the general tensor program formulation proposed in \cite{yang2019wide, yang2021tensor}; to non-Gaussian laws for the random parameters, such as discrete cases or heavy-tailed distributions like stable laws where the Gaussian CLT fails \cite{peluchetti_stable_2020}, by using finer probabilistic tools, such as quantitative multi-dimensional CLT's in the Wasserstein space \cite{bonis_steins_2020}.

Another interesting question is whether \eqref{eq:thesis-clean} are sharp in some sense (eventually allowing for general non Gaussian distributions in the weights and biases parameters) and what are the properties, e.g.\ regularity, of the optimal transport map  pushing the Gaussian law to the distribution of the output of the neural network, whose existence follows from  general results  
\cite[theorem 9.4]{villani2009optimal}: more precise information may help to transfer finer results from Gaussian processes to neural networks. 

A further direction concerns the validity of similar quantitative approximations after the training procedure, e.g.\ by coupling it with the gradient flow description provided by the already mentioned  Neural Tangent Kernel theory. Alternatively, one could investigate if the arguments in \cite{hron_exact_2020} for the exact  Bayes posterior may be combined with our quantitative bounds.

\subsection{Structure of the paper} Section~\ref{sec:notation} is devoted to introduce notation and basic facts about neural networks, Gaussian processes and the Wasserstein distance. Section~\ref{sec:proof}  contains a more precise statement of our main result, Theorem~\ref{thm:main-induction} and its entire proof. Section~\ref{sec:lemma} gives with the proof of technical Lemma~\ref{lem:square-root}, while Section~\ref{sec:functional} contains some remarks on the functional limit (i.e., for infinite input sets). Finally, in Section~\ref{sec:numerics} we report the outcomes of some numerical experiments about the sharpness of our main result.

\section{Notation}\label{sec:notation}

Given a set $S$, we write $\R^{S}$ for the set of  real valued functions $f:S \to \R$, and write  $s \mapsto f(s)$, but also frequently use the notation $f[s] = f_s$, to avoid subscripts  (in vectors and matrices). For a subset $T \subseteq S$, we denote by $f[T] = (f[t])_{t \in T} \in \R^{T}$ its restriction to $T$. When $S = \cur{1, 2, \ldots, k}$, then we simply write $\R^S = \R^k$, which is the set of $k$-dimensional vectors $v = (v[i])_{i=1}^k$, while if $S = \cur{1, \ldots, h} \times \cur{1, \ldots, k}$ then we write $\R^S = \R^{h\times k}$, the set of matrices $A = (A[i,j])_{i=1, \ldots, h}^{j=1, \ldots, k}$. The space $\R^S$ enjoys a natural structure of vector space with pointwise sum $(f+g)[s] = f[s]+g[s]$ and product $(\lambda f)[s] = \lambda f[s]$ and a canonical basis $(e_s)_{s \in S}$ with $e_s[t] =  1$ if $s=t$ and $e_s[t]=0$ otherwise. 

Given sets $S$, $T$, the tensor product operation between $f \in \R^S$ and $g \in \R^T$ defines $f \otimes g \in \R^{S\times T}$ with $(f \otimes g)[s,t] = f(s) g(t)$. The canonical basis for $\R^{S\times T}$ can be then written as $(e_{(s,t)} = e_s\otimes e_t)_{s\in S, t\in T}$.  If $S$, $T$ are finite, then any $A \in \R^{S \times T}$ can be identified also with the space of linear transformations from $\R^{T}$ to $\R^S$ via the usual matrix product
$$ A: f \mapsto Af \quad \text{where} \quad (Af)[s] = \sum_{t \in T} A[s,t] f[t].$$
The identity matrix $\Id_S \in \R^{S\times S}$ is defined as $\Id_S = \sum_{s} e_s \otimes e_s$, i.e., $\Id_S[s,t] = e_s[t]$, and write $\Id_{k}$ or $\Id_{h\times k}$ in the case of $\R^k$ or $\R^{h \times k}$ respectively.
If $S = S_1 \times S_2$ and $T= T_1 \times T_2$ and $A \in \R^{S_1\times T_1}$, $B \in \R^{S_2 \times T_2}$, then, up to viewing $A \otimes B \in \R^{(S_1\times T_1)\times (S_2 \times T_2)}$ as an element in $\R^{(S_1\times S_2) \times (T_1 \times T_2)}$, the following identity holds:
$$ (A \otimes B) (f \otimes g) = (A f) \otimes (B g)$$
for every $f \in \R^{T_1}$, $g \in \R^{T_2}$. When $S=T$, then any $A \in \R^{S \times S}$ can be also seen as a quadratic form over $\R^S$: $A$ is symmetric if $A$ equals its transpose $A^\tau[s,t] = A[t,s]$, i.e., $A[s,t]=A[t,s]$ for $s$, $t \in S$ and positive semidefinite if, for every finite subset $T \subseteq S$ the restriction of $A$ to $T\times T$, that we denote by $A[T]$ (instead of $A[T \times T]$) is positive semidefinite as a matrix, i.e., for every $v \in \R^T$, 
$$ \sum_{s,t \in T} v[s] A[s,t] v[t] \ge 0.$$
If $S$ is finite, we define the trace of $A \in \R^{S\times S}$ as $\operatorname{tr}(A) = \sum_{s\in S} A[s,s]$. We  use this notion to define a scalar product on $\R^S$ as $$ \ang{f, g} = \operatorname{tr}(f \otimes g) = \sum_{s\in S} f[s]g[s],$$
which induces the norm
$$\nor{f} =\sqrt{ \operatorname{tr}( f \otimes f)} = \sqrt{ \sum_{s \in S} |f[s]|^2},$$ yielding the Euclidean norm in case of a vectors in $\R^k$ and the Frobenius (or Hilbert-Schmidt) norm for  matrices $\R^{h\times k}$. The tensor product is associative and the norm is multiplicative, i.e., $\nor{f\otimes g} = \nor{f}\nor{g}$.

Given a symmetric positive semi-definite $A\in \R^{S \times S}$ with $S$ finite, its square root $\sqrt{A}$ is defined via spectral theorem as $\sqrt{A} = U \sqrt{D} U^\tau$ where $U \in \R^{S\times S}$ is orthogonal, i.e., $U U^\tau = \Id$ and $U^\tau A U = D= \sum_{s} d_s e_s \otimes e_s$ is diagonal with $d_s \ge 0$ (the eigenvalues of $A$) and $\sqrt{D} = \sum_s \sqrt{d_s}e_s \otimes e_s$ is the diagonal with entries being the square roots of the entries of $D$. If $A$, $B \in \R^{S \times S}$ are symmetric and positive semi-definite, then \cite[Proposition 3.2]{van1980inequality} gives
\begin{equation}\label{eq:ando}
\nor{ \sqrt{A} - \sqrt{B}} \le \frac{1}{\sqrt{ \lambda(A) } } \nor{ A - B},
\end{equation}
 where $\lambda(A) = \min_s d_s$ denotes its smallest eigenvalue (possibly zero: in such a case we interpret the right hand side as $+\infty$).  Finally, we will use the fact that $\sqrt{A \otimes B} = \sqrt{A} \otimes \sqrt{B}$ if $A \in \R^{S\times S}$, $B\in \R^{T \times T}$ are both symmetric and positive-semidefinite, by interpreting the identity in $\R^{(S \times T)\times (S\times T)}$. 
 

\subsection{Random variables and Gaussian processes} For a random variable $X$ with values on $\R^S$, write $\EE\sqa{X} \in \R^S$ for its mean value, i.e., $\EE\sqa{X}[s] = \EE\sqa{X[s]}$. The second moment of $X$ is defined as $\EE\sqa{ X \otimes X} \in \R^{S\times S}$, i.e., the collection $(\EE\sqa{X[s] X[t]})_{s,t\in S}$, which is symmetric and positive semi-definite.  We always tacitly assume that these are well-defined. The covariance matrix $\Sigma(X) \in \R^{S\times S}$ of $X$ is defined as the second moment of the centered variable $X-\EE\sqa{X}$, which coincides with its second moment if $X$ is centered, i.e., $\EE\sqa{X} =0$. Recalling that $\nor{X}^2 = \tr(X \otimes X)$, and exchanging expectation and the trace operation, we have that $\EE\sqa{ \nor{X}^2} = \tr( \EE\sqa{X \otimes X} )$, hence
\begin{equation}\label{eq:trace-norm} \EE\sqa{\nor{X}^2 } = \tr(\Sigma(X)), \quad \text{
if $X$ is centered.}\end{equation}
If $A$ is a random variable with values in $\R^{T \times S}$, its second moment $$\EE\sqa{A\otimes A} \in \R^{(T\times S) \times (T \times S)}$$ can be viewed also as in $\R^{(T\times T) \times (S\times S)}$, which is a useful point of view we often adopt. 

Given a symmetric positive semi-definite $K \in \R^{S \times S}$, we write $\mathcal{N}(K)$ for the centered Gaussian distribution on $\R^S$ with covariance (or, equivalently, second moment) $K$, i.e., the law of any Gaussian random variable $X$ with values in $\R^S$, $\EE\sqa{X} = 0$ and $\Sigma(X) = K$. If $S$ is infinite, the law is Gaussian if for every finite subset $T \subseteq S$, the restriction $X[T] = (X[t])_{t \in T}$ has a multivariate Gaussian law with covariance $K[T] = K[T\times T]$. The variable $X = (X[s])_{s \in S}$ is also called a Gaussian process (or field) over $S$.



\subsection{Wasserstein distance}\label{sec:wasserstein}  We recall without proof basic facts about the Wasserstein distance, referring to any monograph such as \cite{villani2009optimal, ambrosio2014metric, peyre2019computational, santambrogio2015optimal} for details.  Let $S$ be  finite. The quadratic Wasserstein distance between  two probability measures $p$, $q$ on $\R^S$ is defined  as
\[ \W_2(p, q) = \inf\cur{ \sqrt{\EE\sqa{ \nor{X-Y}^2}} \, :  \text{$X$, $Y$ random variables with $\mathbb{P}_X = p$, $\mathbb{P}_Y = q$}},\]
where the infimum runs over all the random variables $X$, $Y$ (jointly defined on some probability space) with marginal laws  $p = \mathbb{P}_X$ and $q=\mathbb{P}_Y$. 
For simplicity, we allow for a slight abuse of notation and often write $\W_2(X,Y)$ instead of $\W_2(\mathbb{P}_X, \mathbb{P}_Y)$, when we consider two random variables $X$, $Y$ with values on $\R^S$. 
In particular, it holds
\begin{equation}\label{eq:w-2-trivial-bound} \W_2(X,Y) \le \sqrt{\EE\sqa{ \nor{X-Y}^2}}.\end{equation}

With this notation, a  sequence of random variables $(X_n)_n$ converges towards a random variable $X$, i.e., $\lim_{n \to \infty}\W_2(X_n, X) = 0$ if and only if $\lim_{n \to \infty} X_n$ in law together with their second moments $\lim_{n \to \infty} \EE\sqa{X_n \otimes X_n}  = \EE\sqa{X\otimes X}$. 
Given $X$, $Y$,$Z$ all with values on $\R^S$, the triangle inequality holds
\begin{equation}\label{eq:triangle}
\W_2(X,Z) \le \W_2(X,Y)+\W_2(Y,Z),
\end{equation}
and, if $Z$ is independent of $X$ and $Y$,  the following inequality holds:
\begin{equation}\label{eq:sum} \W_2(X+Z, Y+Z) \le \W_2(X, Y).\end{equation}

The distance squared enjoys a useful convexity property, which we can state as follows: 
given random variables $X$, $Y$, with values on $\R^S$ and $Z$, taking values on some $\R^T$, then
\begin{equation}\label{eq:convexity} \W_2^2( X, Y) \le  \int_{\R^T} \W_2^2( \mathbb{P}_{X|Z=z} , \mathbb{P}_{Y}) d \mathbb{P}_Z(z).\end{equation}

Finally, if $X$, $Y$ are centered Gaussian random variables with values on $\R^S$ ($S$ finite), it holds 
\begin{equation}\label{eq:gaussian} \W_2(X, Y) \le \nor{ \sqrt{\Sigma(X)} - \sqrt{\Sigma(Y)}},\end{equation}
which can be easily seen by considering a standard Gaussian variable $Z$ on $\R^S$, i.e., centered with $\Sigma(Z) = \Id_{S}$ and letting $X = \sqrt{\Sigma(X)} Z$ and $Y = \sqrt{\Sigma(Y)}Z$.

\subsection{Deep Neural Networks}\label{sec:nn}
We consider only the so-called fully connected (or multi-layer perceptron) architecture, which consists of a sequence of hidden layers stacked between the input and output layers, where each node  is connected with all nodes in the subsequent layer. Let $L \ge 1$ and let ${\bf n} = (n_0, n_1, \ldots, n_L)$ denote the size of the layers, so that the input belongs to $\R^{n_0}$ and the overall network outputs a vector in $\R^{n_{L}}$. Given an activation function $\sigma: \R \to \R$ and a set of weights and biases (in this section they do not need to be random variables)
$${\bf W} = (W^{(0)}, W^{(1)}, \ldots, W^{(L-1)}), \quad  {\bf b} = (b^{(0)}, b^{(1)}, \ldots, b^{(L-1)}),$$
where, for every $\ell =0, \ldots, L-1$, one has $W^{(\ell)} \in \R^{n_{\ell+1} \times n_{\ell}}$ and $b^{(\ell)} \in \R^{n_{\ell+1}}$, we define 
$$ f^{(1)} : \R^{n_0} \to \R^{n_1}, \quad f^{(1)}[x] = W^{(0)}x+ b^{(0)},$$
and recursively, for $\ell=2, \ldots, L$,
$$ f^{(\ell)} : \R^{n_0} \to \R^{n_{\ell}}, \quad f^{(\ell)}[x] = W^{(\ell-1)}\sigma(f^{(\ell-1)}[x]) + b^{(\ell -1)},$$
where the composition  $\sigma$ will be always understood componentwise. Notice that $f^{(\ell)}$ is a function of weights and biases $(W^{(i)}, b^{(i)})_{i=0}^{\ell-1}$. Given a finite subset of inputs $\X = \cur{x_i}_{i=1,\ldots, k} \subseteq \R^{n_0}$, we identify it with a slight abuse of notation with a matrix $\X = \sum_{i=1}^k x_i \otimes e_i \in \R^{n_0 \times k}$ and consider 
the restrictions $f^{(\ell)}[\X]$ that we interpret also as matrices 
$$ f^{(\ell)}[\X] =\sum_{i=1}^k f^{(\ell)}[x_i] \otimes e_{i} \in \R^{n_{\ell} \times k}.$$
We notice that
\begin{eqnarray}  &f^{(1)}[\X] = (W^{(0)} \otimes \Id_k ) \X+ b^{(0)}\otimes 1_k,\label{eq:f-base}\\  & f^{(\ell)}[\X]  = (W^{(\ell-1)} \otimes \Id_k)  \sigma ( f^{(\ell-1)}[ \X] )  + b^{(\ell-1)}\otimes 1_k, \label{eq:f-induction}
 \end{eqnarray}
for $\ell = 2, \ldots, L$, where $1_k  = \sum_{i=1}^k e_i \in \R^k$ denotes the vector with all entries equal to $1$.

\subsection{Neural Network Gaussian Processes} \label{sec:nngp}

The Gaussian process associated to a wide random neural network \cite{lee_deep_2018} is defined merenly in terms of its covariance operator (it is centered) and ultimately depends only on the choice of the activation function $\sigma$, the non-negative weights and biases variances 
$$ \bra{c^{(0)}_w, c^{(0)}_b, c^{(1)}_w, c^{(1)}_b, \ldots, c^{(L-1)}_w, c^{(L-1)}_b} \in \R_+^{2L}$$
as well as the layer dimensions. As with the construction of the network, the definitions are recursive. We first define the kernel $K^{(1)} = (K^{(1)}[x,y])_{x, y \in \R^{n_0}}$ as
$$ K^{(1)}[x,y] = \frac{ c^{(0)}_w}{n_0} \ang{x,y} + c_b^{(0)} = \frac{c^{(0)}_w}{n_0} \sum_{i=1}^{n_0} x[i]y[i] + c_b^{(0)}.$$
It is not difficult to check that $K^{(1)}$ is symmetric and positive semi-definite, thus there exists a centered Gaussian process $(G^{(1)}[x])_{x \in \R^{n_0}}$ on $\R^{n_0}$ with covariance kernel $K^{(1)}$, i.e.,
$$ \EE\sqa{ G^{(1)}[x] G^{(1)}[y]} = K^{(1)}[x,y].$$
For $\ell =2, \ldots, L$, having already defined the positive semi-definite kernel $K^{(\ell-1)} = (K^{(\ell-1)}[x,y])_{x, y \in \R^{n_0}}$, we denote with $(G^{(\ell-1)}[x])_{x \in \R^{n_0}}$ a centered Gaussian process with covariance kernel $K^{(\ell-1)}$ and use it to define
$$ K^{(\ell)}[x,y] = c^{(\ell-1)}_w \EE\sqa{ \sigma(G^{(\ell-1)}[x]) \sigma(G^{(\ell-1)}[y])} + c_b^{(\ell-1)},$$
which can be easily seen to be positive semi-definite. For later use, let us also introduce the kernel
\begin{equation}\label{eq:k-ell-0} K^{(\ell)}_0[x,y] = c^{(\ell-1)}_w \EE\sqa{ \sigma(G^{(\ell-1)}[x]) \sigma(G^{(\ell-1)}[y])}.\end{equation}
Finally, given layer dimensions ${\bf n} = (n_0, n_1, \ldots, n_L)$, the Gaussian process approximating the layer output $f^{(\ell)}[x]$, is defined as $n_\ell$ independent copies of $G^{(\ell)}$, i.e., a Gaussian process on $S = \R^{n_0} \times \cur{1, \ldots, n_\ell}$, whose covariance kernel is then $ K^{(\ell)}\otimes \Id_{n_\ell}$. To keep notation simple, at the cost of a slight ambiguity of the notation, we also write $G^{(\ell)}[x] = (G^{(\ell)}[x,i])_{i=1,\ldots, n_\ell} \in \R^{n_\ell}$ for the (vector-valued) Gaussian process. In analogy with the notation for neural networks, given $k$ inputs $\X = \cur{x_i}_{i=1}^k$, write then $G^{(\ell)}[\X] = \sum_{i=1}^k G^{(\ell)}[x_i] \otimes e_i \in \R^{n_{\ell} \times k}$, which is a centered Gaussian random variable with covariance
$$ \Sigma\bra{G^{(\ell)}[\X]} =  \Id_{n_\ell} \otimes K^{(\ell)}[\X] \in \R^{(n_\ell \times n_\ell)\times (k \times k)},$$
where $K^{(\ell)}[\X] = (K^{(\ell)}[x,y])_{x, y \in \X}$.





\section{Proof of Main Result}\label{sec:proof}

We consider the fully connected architecture described in the previous section, where weights ${\bf W}$ and biases ${\bf b}$ are all independent Gaussian random variables, centered with variances
$$ \Sigma(W^{(\ell)}) = \frac{c^{(\ell)}_w}{n_{\ell}} \Id_{n_{\ell+1} \times n_{\ell}}, \quad \Sigma(b^{(\ell)}) = c^{(\ell)}_b \Id_{n_{\ell+1}},$$
where $(c^{(\ell)}_w, c^{(\ell)}_b)_{\ell=0}^{L-1} \in \R_+^{2L}$. We also assume that the activation function $\sigma$ is Lipschitz, i.e., for some finite constant $\Lip(\sigma)$
\begin{equation}\label{eq:lip-sigma} |\sigma(z) - \sigma(z')| \le \Lip(\sigma) |z-z'| \quad \text{for all $z$, $z' \in \R$.}\end{equation}
This is the case with $\Lip(\sigma)=1$ for the common choice of ReLU activation $\sigma(z) = z^+$.

We state and prove a more precise version of our main result. 

\begin{theorem}\label{thm:main-induction}
With the above notation for the deep neural network outputs $\cur{ f^{(\ell)}[\X]}_{\ell=1}^L$ with random weights and biases, and the associated Gaussian processes $\cur{G^{(\ell)}[\X]}_{\ell=2}^L$ evaluated at $k$ inputs $\X = \cur{x_i}_{i=1}^k \subseteq \R^{n_0}$, there exists positive finite constants $(C^{(\ell)})_{\ell=1}^L$ such that, for every $\ell = 1, \ldots, L$, it holds
\begin{equation}\label{eq:induction-thesis} \W_2\bra{ f^{(\ell)}[\X], G^{(\ell)}[\X]} \le \sqrt{n_\ell} \sum_{i=1}^{\ell-1} \frac{C^{(i+1)}\Lip(\sigma)^{\ell-1-i} \sqrt{ \prod_{j=i+1}^{\ell-1} c^{(j)}_w }}{\sqrt{n_i}}.\end{equation}
Each $C^{(\ell)}$ depends upon $\sigma$, $\X$ and only the weights $(c^{(i)}_{w}, c^{(i)}_{b})_{i=0}^{\ell}$.
\end{theorem}


\begin{remark}\label{rem:deep-limit}
Inequality \eqref{eq:induction-thesis} is evidently more precise than \eqref{eq:thesis-clean}. In particular, besides recovering the Gaussian convergence of the network outputs (for a finite input set) in the wide limit, i.e., if all $(n_i)_{i=1}^{L-1}$ become large, it allows for some partial consideration in the deep limit, i.e., if $L$ becomes large and all the $n_i$'s are uniformly bounded.  Let $\ell=L \to \infty$ in \eqref{eq:induction-thesis} and assume e.g.\ that $\Lip(\sigma)\le 1$ (as in the ReLu case) and all weights variances are constants, $c_w^{(i)} = c_w$. The $i$-th term in the sum is then multiplied by the exponential factor $\sqrt{ c_w^{L-i-2}}$ which becomes negligible as $L\to \infty$, if $c_w<1$. According to  \cite[Section 3.4]{roberts2022principles}, the ``critical'' case in the deep narrow limit is indeed the choice $c_w= 1$, where the network behaviour is ``chaotic'' (and not Gaussian), and indeed our inequality becomes ineffective. 
\end{remark}

Before we provide the proof of Theorem~\ref{thm:main-induction}, which is performed by induction over $\ell=1, \ldots, L$, we inform the reades we extensively use tensor product notation. This is necessary to take into account the joint distributions of the various outputs, which are independent only in the Gaussian case (or in the wide network limit). The  interested reader that would prefer to avoid tensor products but still have a good understanding of the argument is strongly invited to specialize our derivation to the  case of a single input and a one-dimensional output.

\noindent \emph{Base case.} The case $\ell =1$ is straightforward, since $f^{(1)}[\X]$ is a linear function of the Gaussian variable $W^{(0)}$ and  $b^{(0)}$, thus it has Gaussian law, centered with covariance
\[ \begin{split} \Sigma \bra{f^{(1)}[\X]} & = \Sigma \bra{(W^{(0)} \otimes \Id_k ) \X  + b^{(0)}\otimes 1_k} \\
& = \Sigma \bra{(W^{(0)} \otimes \Id_k ) \X } + \Sigma\bra{b^{(0)}\otimes  1_k} \quad \text{by independence,}\\
& =  \Id_{n_1} \otimes K^{(1)}[\X, \X],
\end{split}\]
where in the last step we used the following computation  which we state in a general form for later use (here we used $A=\X$ and $\ell=0$). 

\begin{lemma}
For any $\ell = 0, \ldots, L-1$, given $A \in \R^{n_\ell \times k}$, one has
\begin{equation}\label{eq:sigma-WA} \Sigma( (W^{(\ell)}\otimes \Id_k) A) = \Id_{n_{\ell+1} \times n_{\ell+1}} \otimes \bar{A},\end{equation}
where  $\bar{A}[j_1,j_2] = \frac {c^{(\ell)}} {n_\ell} \sum_{m=1}^{n_\ell} A[m, j_1] A[m,j_2]$ and
\begin{equation}\label{eq:mean-norm-WA} \EE\sqa{ \nor{ (W^{(\ell)}\otimes \Id_k) A }^2} = \frac{ c^{(\ell)} n_{\ell+1}}{n_\ell} \nor{A}^2.
\end{equation}

\end{lemma}


\begin{proof}
The first identity is an explicit computation, using the covariance of $W^{(\ell)}$:
\[\begin{split} \Sigma( (W^{(\ell)}\otimes \Id_k) A)[i_1, j_1, i_2, j_2] & = \EE\sqa{ \bra{ (W^{(\ell)}\otimes \Id_k) A }[i_1,j_1] \bra{(W^{(\ell)}\otimes \Id_k) A} [i_2,j_2]}\\
 & =  \sum_{m_1,m_2=1}^{n_\ell} \EE\sqa{ W^{(\ell)}[i_1, m_1] A [m_1,j_1] W^{(\ell)}[i_2, m_2]  A [m_2,j_2]}\\
 & = e_{i_1}[i_2] \frac {c^{(\ell)}} {n_\ell} \sum_{m=1}^{n_\ell} A[m, j_1] A[m,j_2].
\end{split}\]
The second identity follows from  \eqref{eq:trace-norm}.
\end{proof}

\noindent \emph{Induction step.}  We assume that \eqref{eq:induction-thesis} holds for some $1 \le \ell \le L-1$ and prove it for $\ell+1$. To simplify the notation, we omit to write $[\X]$ in what follows, and consider any probability space where random variables with the same laws as $f^{(\ell)}[\X]$ and $ G^{(\ell)}[\X]$ are jointly defined (but we assume nothing on their joint law). We denote them as $f^{(\ell)}$ and $G^{(\ell)}$.

Without loss of generality, we assume that the weights $W^{(\ell)}$ and biases $b^{(\ell)}$ are defined on the same space and independent of $f^{(\ell)}$ and $G^{(\ell)}$, otherwise we enlarge it to define independent copies of $W^{(\ell)}$ and $b^{(\ell)}$. We define auxiliary random variables
$$ h^{(\ell+1)} = (W^{(\ell)} \otimes \Id_k) \sigma\bra{ G^{(\ell)}}, \quad g^{(\ell+1)} = h^{(\ell+1)}+b^{(\ell)}\otimes 1_k.$$
To take into account the biases, which play practically no role in our bounds, we also assume that a centered Gaussian random variable $H^{(\ell+1)}$ is defined on the same space, with values in $\R^{n_{\ell+1}\times k}$  and covariance
$$  \Id_{n_{\ell+1}} \otimes  K^{(\ell+1)}_0[\X \times \X] \in \mathbb{R}^{(n_{\ell+1}\times n_{\ell+1}) \times (k\times k)},$$
with the kernel $K^{(\ell+1)}_0$ introduced in \eqref{eq:k-ell-0}. 
%
Without loss of generality, we also assume that $H^{(\ell+1)}$ is independent from $b^{(\ell)}$, so that
$$ H^{(\ell+1)}+ b^{(\ell)}\otimes 1_k$$
is Gaussian with the same law as $G^{(\ell+1)} = G^{(\ell+1)}[\X]$.

By the triangle inequality, we now split
$$  \W_2\bra{ f^{(\ell+1)}, G^{(\ell+1)}} \le  \W_2\bra{ f^{(\ell+1)}, g^{(\ell+1)}} + \W_2\bra{ g^{(\ell+1)}, G^{(\ell+1)}},$$
and bound separately the two terms.

\noindent \emph{First term. } We prove that
\begin{equation}
\label{eq:first-term}
\W_2^2\bra{ f^{(\ell+1)}, g^{(\ell+1)}} \le \frac{ n_{\ell+1}}{n_{\ell}} c^{(\ell)} \Lip(\sigma)^2 \EE\sqa{ \nor{   f^{(\ell)}-G^{(\ell)}}^2}.
\end{equation}
Indeed,
\[ \begin{split} \W_2^2\bra{ f^{(\ell+1)}, g^{(\ell+1)}} & =  \W_2^2\bra{W^{(\ell)}\sigma( f^{(\ell)})+ b^{(\ell)}, W^{(\ell)} \sigma( G^{(\ell)}) + b^{(\ell)}}\\
& \le  \W_2^2\bra{W^{(\ell)}\sigma( f^{(\ell)}), W^{(\ell)} \sigma( G^{(\ell)}) } \quad \text{by \eqref{eq:sum},}\\
& \le \EE\sqa{ \nor{ W^{(\ell)}\sigma( f^{(\ell)})-  W^{(\ell)} \sigma( G^{(\ell)}) }^2 } \quad \text{by \eqref{eq:w-2-trivial-bound}.}
\end{split}\]
To proceed further, we condition upon $f^{(\ell)}$ and $G^{(\ell)}$, so that, with $A = \sigma( f^{(\ell)})-\sigma( G^{(\ell)})$  in \eqref{eq:mean-norm-WA}, we obtain
\[ \EE\sqa{ \nor{ W^{(\ell)}\sigma( f^{(\ell)})-  W^{(\ell)} \sigma( G^{(\ell)}) }^2\,  \Big| \,  f^{(\ell)}, G^{(\ell)}} = \frac{  c^{(\ell)} n_{\ell+1}}{n_{\ell}}  \nor{  \sigma( f^{(\ell)})-\sigma( G^{(\ell)})}^2. \]
Finally, we use \eqref{eq:lip-sigma}, to bound from above
$$  \nor{  \sigma( f^{(\ell)})-\sigma( G^{(\ell)})}^2 \le \Lip(\sigma)^2 \nor{   f^{(\ell)}-G^{(\ell)}}^2,$$
and conclude with \eqref{eq:first-term}.


\noindent \emph{Second term.} We prove that
\begin{equation}\label{eq:second-term}   \W_2^2\bra{ f^{(\ell+1)}, G^{(\ell+1)}} \le  \frac{ n_{\ell+1}}{n_{\ell}}  C^{(\ell+1)}, 
\end{equation}
for a positive finite constant $C^{(\ell+1)}$ depending on $\X$, $\sigma$ and $(c_w^{(i)}, c_b^{(i)})_{i=0}^{\ell}$ only (an explicit expression is \eqref{eq:c-ell} below).

To this aim, we use again \eqref{eq:sum} to remove the bias terms,
\[\begin{split} \W_2\bra{ g^{(\ell+1)}, G^{(\ell+1)}} &= \W_2\bra{ h^{(\ell+1)} + b^{(\ell)}\otimes 1_k, H^{(\ell+1)} + b^{(\ell)}\otimes 1_k}\\
& \le \W_2\bra{ h^{(\ell+1)}, H^{(\ell+1)} }.
\end{split}\]
Next, we use \eqref{eq:convexity} with $X = h^{(\ell+1)}$, $Y = H^{(\ell+1)}$ and $Z=G^{(\ell)}$. 
Indeed, conditioning upon $G^{(\ell)} = z \in \R^{n_{\ell}\times k}$, the random variable
$$h^{(\ell+1)} = (W^{(\ell)} \otimes \Id_k) \sigma\bra{ G^{(\ell)}} = (W^{(\ell)} \otimes \Id_k) \sigma\bra{ z}$$
has centered Gaussian law whose covariance is provided by \eqref{eq:sigma-WA} applied with $A = \sigma(z)$: $\Id_{n_{\ell+1}} \otimes \overline{\sigma(z)}$, where $\overline{\sigma(z)}\in \R^{k \times k}$ is given by
$$ \overline{\sigma(z)} [j_1, j_2] = \frac{ c_w^{(\ell)} }{n_\ell} \sum_{m=1}^{n_\ell}\sigma\bra{ z[m,j_1]}\otimes \sigma\bra{ z[m,j_2]}.$$
By \eqref{eq:gaussian}, we bound from above
\[\begin{split} \W_2^2\bra{ \mathbb{P}_{ h^{(\ell+1)} | G^{(\ell)} = z}, \mathbb{P}_{H^{(\ell+1)}} } & \le \nor{ \sqrt{\Id_{n_{\ell+1}} \otimes  \overline{\sigma(z)}} - \sqrt{ \Id_{n_{\ell+1}} \otimes K^{(\ell+1)}_0 }}^2\\
& = \nor{ \Id_{n_{\ell+1}  } \otimes \bra{ \sqrt{ \overline{\sigma(z)}} -  \sqrt{K^{(\ell+1)}_0} }}^2\\
& = n_{\ell+1} \nor{\sqrt{ \overline{\sigma(z)}} -  \sqrt{K^{(\ell+1)}_0}}^2. \end{split}
\]
Therefore, by \eqref{eq:convexity}, we have
$$ \W_2^2\bra{ h^{(\ell+1)}, H^{(\ell+1)} } \le n_{\ell+1} \EE\sqa{ \nor{\sqrt{ \overline{\sigma( G^{\ell})}} -  \sqrt{K^{(\ell+1)}_0}}^2}.$$

To conclude, we use the following lemma,  whose proof is postponed in the next section.

\begin{lemma}\label{lem:square-root}
Let $X= (X[i])_{i=1}^n$ be independent and identically distributed random variables with values in $\R^{k}$ and finite fourth moment. Assuming that $X$ is not identically null, let $M = \EE\sqa{ X[1]\otimes X[1]}$ and define the following random variable, taking values in $\R^{k\times k}$,
$$ M_n = \frac 1 n \sum_{i=1}^n X[i] \otimes X[i].$$
Then,
$$ \EE\sqa{ \nor{\sqrt{M_n} -  \sqrt{M}}^2 } \le \frac{\EE\sqa{ \nor{ X[1]\otimes X[1] - M}^2}}{n \lambda^+(M)},$$
where $\lambda^+(M)>0$ denotes the smallest strictly positive eigenvalue of $M$ (which exists since $X$, hence $M$, is not identically null). The numerator in the right hand side is finite if $X$ has finite fourth moment.
\end{lemma}

Indeed, defining $X[i] = \sigma(G^{\ell})[i]$, i.e., $X[i,j] = \sigma(G^\ell[i,j])$, the independence assumption is satisfied, because $G^\ell$ is Gaussian centered with covariance $\Id_{n_\ell \times n_\ell} \otimes K^{(\ell)}[\X]$ hence $G^{\ell}[i_1,j_1]$ and $G^{\ell}[i_2,j_2]$ are not correlated if $i_1 \neq i_2$, which yields independence (which is then preserved after composition with $\sigma$). Since $\sigma$ is not identically null, so is $X$ hence the thesis with $n=n_{\ell}$ yields
$$ \EE\sqa{ \nor{\sqrt{\overline{\sigma( G^{\ell})}} -  \sqrt{K^{(\ell+1)}_0}}^2} \le \frac{ (C^{(\ell+1)})^2}{n_\ell},$$
where the finite constant $C^{(\ell+1)}$ is  explicitly
\begin{equation}\label{eq:c-ell} C^{(\ell+1)} = \sqrt{ \EE\sqa{ \nor{ \sigma(G^{\ell})[1]\otimes \sigma(G^{\ell})[1] - K^{(\ell+1)}_0}^2}/\lambda^+(K^{(\ell+1)}_0)},\end{equation}
and depends uniquely on the activation function $\sigma$, the input set $\X$ and the weights variances $(c_{w}^{(i)}, c_{b}^{(i)})_{i=1}^{\ell}$. The proof of \eqref{eq:second-term} is completed.

\noindent \emph{Conclusion.} 
We have proved that
\[  \W_2\bra{ f^{(\ell+1)}, G^{(\ell+1)}} \le    \sqrt{ \frac{ n_{\ell+1}}{n_{\ell}}} \bra{ \Lip(\sigma)  \sqrt{ c^{(\ell)}\EE\sqa{ \nor{ f^{(\ell)}- G^{(\ell)}}^2}} +C^{(\ell+1)} }.\]
Recalling that we are considering any probability space where random variables with the same laws as $f^{(\ell)}[\X]$ and $G^{(\ell)}[\X]$ are defined, we consider the infimum over such spaces in the above inequality and conclude that
\[\begin{split}   \W_2\bra{ f^{(\ell+1)}, G^{(\ell+1)}} \le    \sqrt{ \frac{ n_{\ell+1}}{n_{\ell}}} \bra{ \Lip(\sigma) \sqrt{ c^{(\ell)} }  \W_2\bra{ f^{(\ell)}, G^{(\ell)}} +C^{(\ell+1)} }
\end{split}.\]
Using the inequality provided by the induction assumption we obtain the thesis.

\section{Proof of Lemma \ref{lem:square-root}}\label{sec:lemma}

The argument relies on \eqref{eq:ando} and the fact that, by the i.i.d.\ assumption
$$  M = \EE\sqa{M_n}, \quad \text{and} \quad \Sigma (M_n) = \Sigma\bra{  \frac 1 n \sum_{i=1}^n X[i]\otimes X[i] } =  \frac {\Sigma(X[1]\otimes X[1])} n, $$
so that, by \eqref{eq:trace-norm},
$$\EE\sqa{ \nor{ M_n - M}^2} = \EE\sqa{ \tr( \Sigma(M_n)) } = \frac {\Sigma(X[1]\otimes X[1])} n = \frac{ \EE\sqa{ \nor{ X[1]\otimes X[1] - M}^2}}{n}.$$

However, we need to take into account that  $M$  may not to be strictly positive, hence $\lambda(M)$ in \eqref{eq:ando} could be null. To this aim, we diagonalize $M = U D U^\tau$ with $U$ orthogonal, $U U^\tau = \Id$ and $D = \sum_{i=1}^k d_i e_i \otimes e_i$ diagonal. Assuming without loss of generality that $d_i >0$ if $i \le k'$ and $d_i=0$ otherwise, for some $k' \le k$, we consider the random variables $Y[i] = (U X[i])$, $i=1, \ldots, n$ which are i.i.d.\ with second moment matrix $D$, since
\[\begin{split} \EE\sqa{ Y[1,j_1] Y[1,j_2] } & = \sum_{i_1, i_2=1}^k \EE\sqa{ U[j_1, i_1] X[1,i_1] U[j_2, i_2] X[1,i_2] } \\
& = \sum_{i_1, i_2=1}^k  U[j_1, i_1] M[i_1,i_2] U[j_2, i_2] = (U M U^\tau)[i_1,j_2].\end{split}\]
It follows that $\EE\sqa{ Y[i,j]^2 } =d_j = 0$ if $j > k'$, hence $Y[i,j] =0$ almost surely. Therefore, both $D$ and
$$ D_n = U M_n U^\tau = \frac 1 n \sum_{i=1}^n U (X[i] \otimes X[i])U^\tau = \frac 1 n \sum_{i=1}^n Y[i] \otimes Y[i]$$
have a block structure where only the first $k'\times k'$ entries are not null. It follows that the square root $\sqrt{D_n}$ has the same block structure and by appling \eqref{eq:ando} only to these blocks we obtain
$$ \nor{ \sqrt{D} - \sqrt{D_n}} \le \frac{\nor{ D - D_n}}{\sqrt{ \lambda^+(D)}},$$
where $\lambda^+(D) = \lambda^+(M)$ is strictly  positive.

To conclude, we use that $U$ is orthogonal to obtain that
$$ \sqrt{ M} = U\sqrt{D} U^\tau \quad \text{and} \quad \sqrt{M_n} = U \sqrt{D_n} U^{\tau},$$
as well as the fact that the invariance of the norm $\nor{ U A U^\tau} = \nor{A}$ (for every $A \in \R^{k \times k}$. Thus,
$$ \nor{ \sqrt{M} - \sqrt{M_n}} = \nor{ \sqrt{D} - \sqrt{D_n} } \le \frac{\nor{ M - M_n}}{\sqrt{ \lambda^+(M)}},$$
squaring both sides and taking expectation the thesis follows.

\section{Some remarks on the functional limit}\label{sec:functional}

 A natural question is to let $k\to \infty$ in \eqref{eq:thesis-clean}, or to consider more generally  the neural network (and the Gaussian process) as random variables with values in a (infinite dimensional) space of functions, such as a Hilbert space, e.g., a Lebesgue space $L^2(\X, \mu)$ or even a Banach space, e.g., the space of continuous functions $\mathcal{C}^0(\X)$, for a compact subset $\X \subseteq \R^{n_0}$ and a finite measure $\mu$ on $\X$, e.g., $\X = \mathbb{S}^{n_0-1}$ and $\mu$ the surface measure. The main obstacle appears to be Lemma~\ref{lem:square-root}, where we use the finite dimensionality assumption on $M$ to obtain a spectral gap, i.e., a bound from below for the smallest strictly positive eigenvalue. Indeed, in infinite dimensions the general bound \eqref{eq:ando} still holds but we expect than strictly positive eigenvalues of $M$ accumulate towards $0$ ($M$ becomes a compact operator). A strategy  would be to decompose the action of $M$ into two eigenspaces (corresponding respectively to large and small eigenvalues) and use \eqref{eq:ando} in the former space and H\"older regularity for the square root in the latter (which gives a worse bound). An alternative strategy would be to rely on the uniform regularity bounds for the process provided by Kolmogorov continuity theorem: arguing as in \cite{bracale_large-width_2020}, one may prove that, for every compact $\X \subseteq \R^{n_0}$, and $\gamma \in (0,1)$, letting $\mathcal{C}^{\gamma}(\X)$ be the space of $\gamma$-H\"older continuous functions (with the usual norm $\nor{\cdot}_{\mathcal{C}^\gamma}$), then
$$ \sup_{(n_i)_{i=1}^{L-1}} \EE\sqa{ \nor{ f^{(L)}}_{\mathcal{C}^\gamma}^2 } \le C(\X)$$
for some constant $C(\X)$ depending on  $\X$ (possibly also $\gamma$, $n_L$  and weights and biases variances). A similar bound holds for the Gaussian process $G^{(L)}$. Given any $\varepsilon>0$, one can find a finite set $\mathcal{K} = \cur{x_i}_{i=1}^{K}$  such that $\inf_{x\in K} \nor{x-y} \le \varepsilon$ for every $y \in \X$. Then, the triangle inequality gives
$$ \nor{ f^{(L)} - G^{(L)} }_{\mathcal{C}^0} \le \bra{ \nor{ f^{(L)}}_{\mathcal{C}^{\gamma}} + \nor{ G^{(L)}}_{\mathcal{C}^{\gamma}}}\varepsilon^{\gamma} + \nor{ f^{(L)}[\mathcal{K}] - G^{(L)}[\mathcal{K}] }.$$
If one could jointly define $f^{(L)}$ and $G^{(L)}$ in such a way that their restrictions on the finite set $\mathcal{K}$ satisfy \eqref{eq:thesis-clean} (with $\mathcal{K}$ instead of $\X$), it would imply an inequality of the type
$$ \W_2(  f^{(L)}[\X],  G^{(L)}[\X] ) \le \inf_{\varepsilon>0}\cur{   C(\X) \varepsilon^{\gamma} + C(\mathcal{K}) \sqrt{n_L} \sum_{\ell=1}^{L-1} \frac{1}{\sqrt{n_{\ell}}}},$$
where the first constant $C(\X)$ in the right hand side is independent of $\varepsilon$, while $C(\mathcal{K})$ depends on $\varepsilon$ through $\mathcal{K}$. The Wasserstein distance $\W_2$ above is between probability distributions in $\mathcal{C}^{0}(\X)$ with respect to the cost given by the $\sup$-norm $\nor{\cdot}_{\mathcal{C}^{0}}$. Again, the main issue is that if $\varepsilon\to 0$ then the set $\mathcal{K}$ becomes large and we expect that the constant $C(\mathcal{K})$ to diverge.   It thus appears achievable, although quite technical, to derive uniform bounds on compact sets for the network process and its associated processes.
 
 We finally mention the work \cite{Matsubara2020TheRP}, where quantitative bounds have been obtained  by considering a non-Gaussian (ridgelet) prior distribution on the network weights, that approximates the  Gaussian process in the output space of the network of a shallow Bayesian Neural Network. It is proved there that any Gaussian Process whose covariance regular enough can be approximated by a Bayesian Neural Network.



\section{Numerical Experiments}\label{sec:numerics}
\begin{figure}[h] 
    \centering
    \includegraphics[width=.45\textwidth]{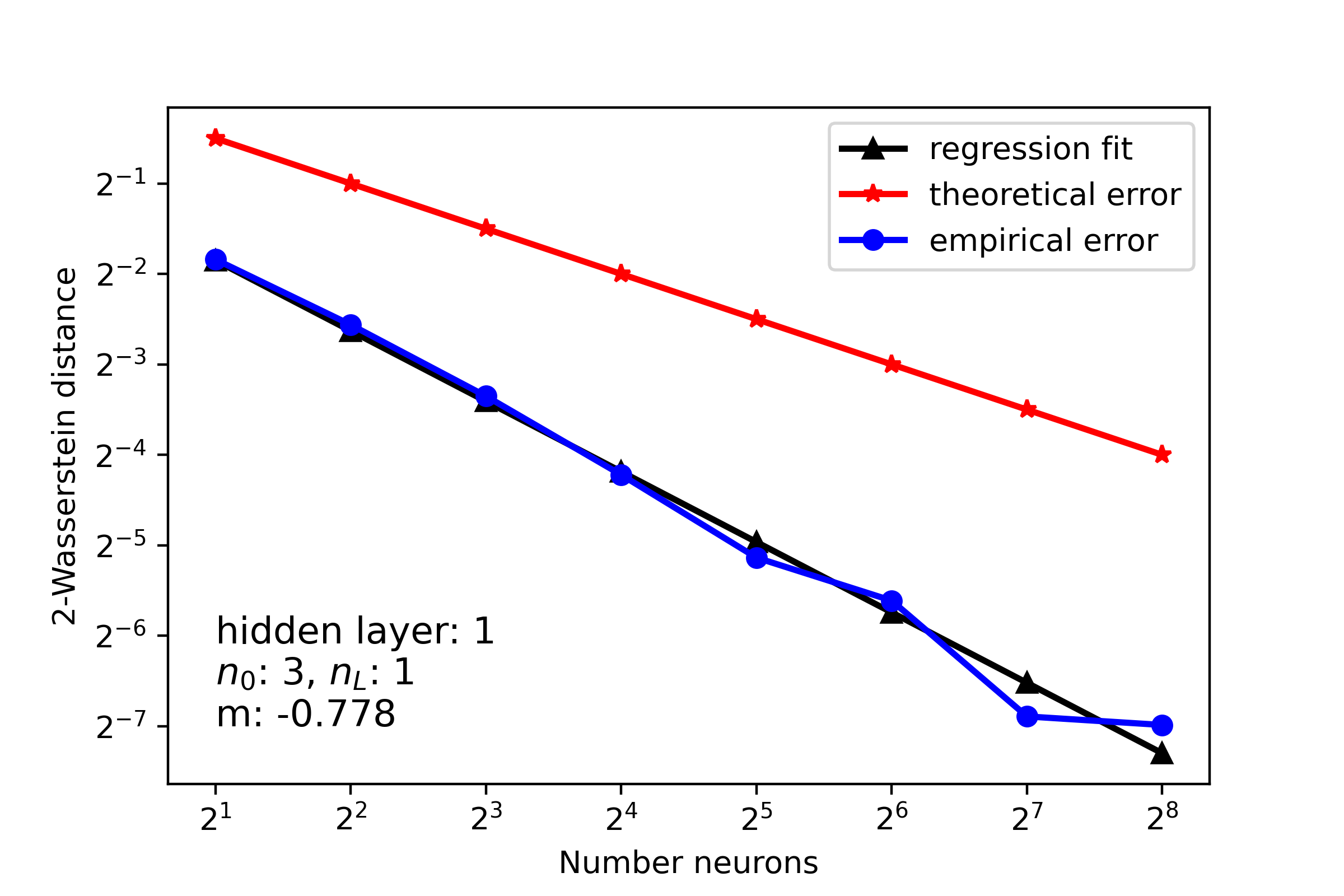}
    \includegraphics[width=.45\textwidth]{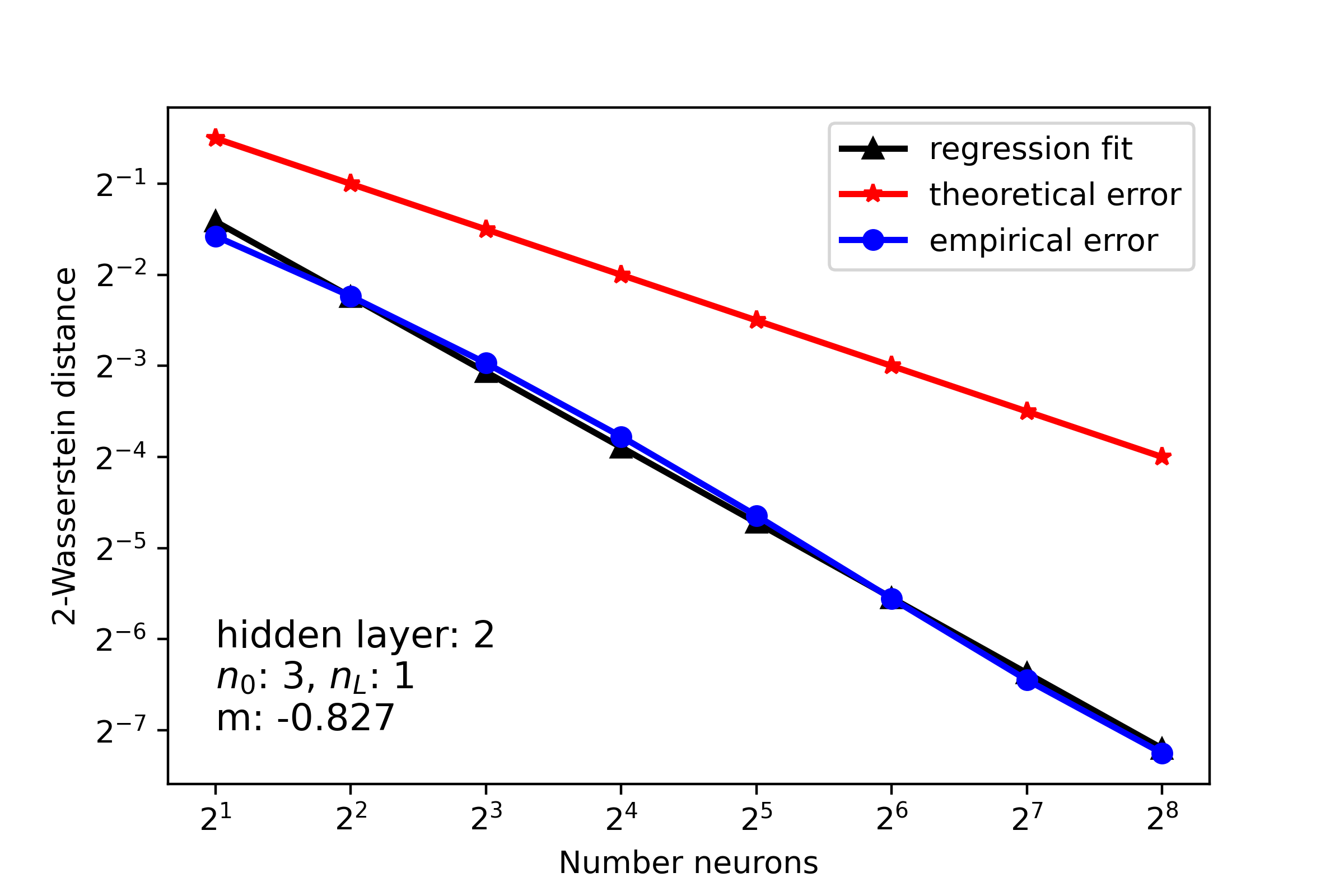}
    \includegraphics[width=.45\textwidth]{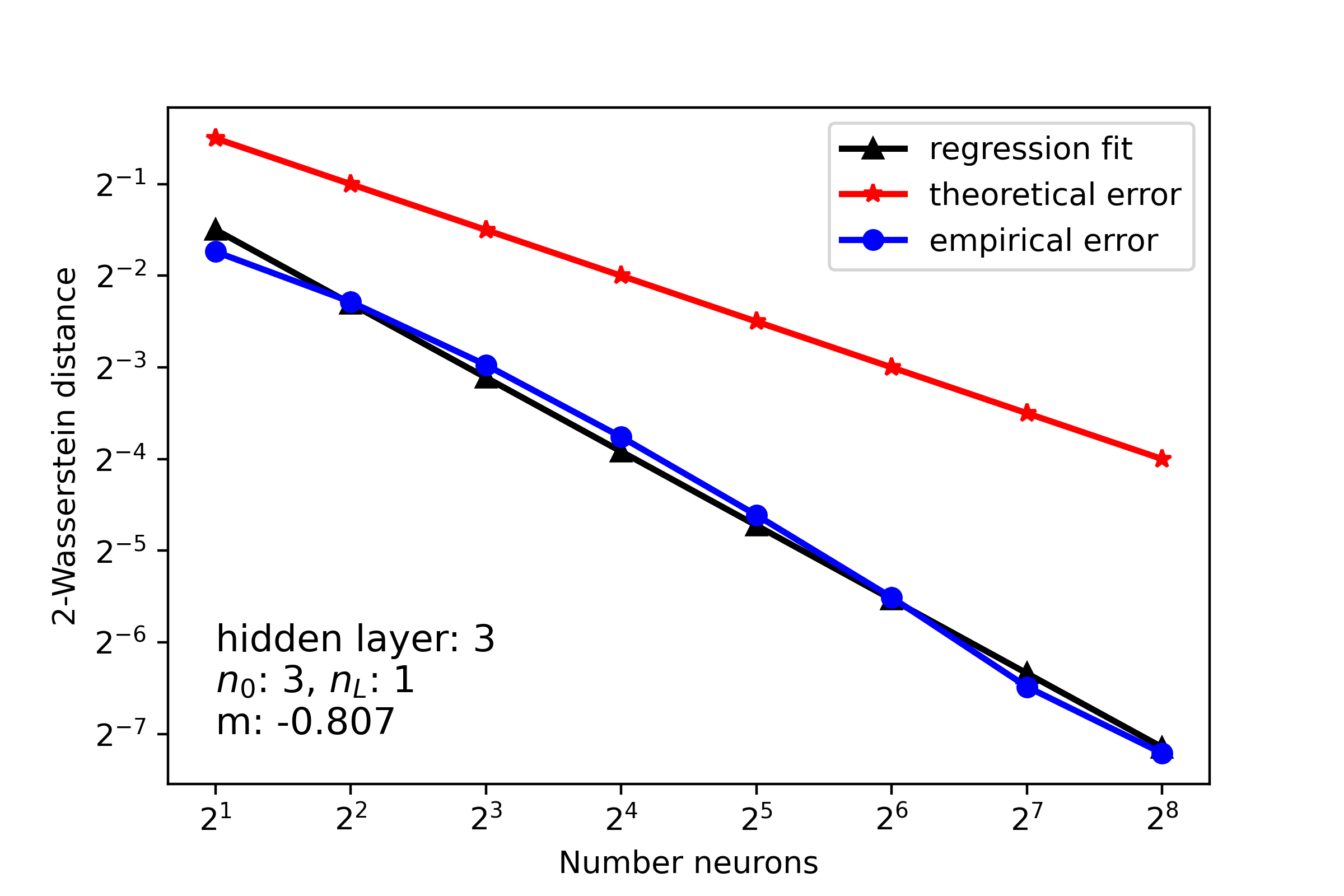}
    \includegraphics[width=.45\textwidth]{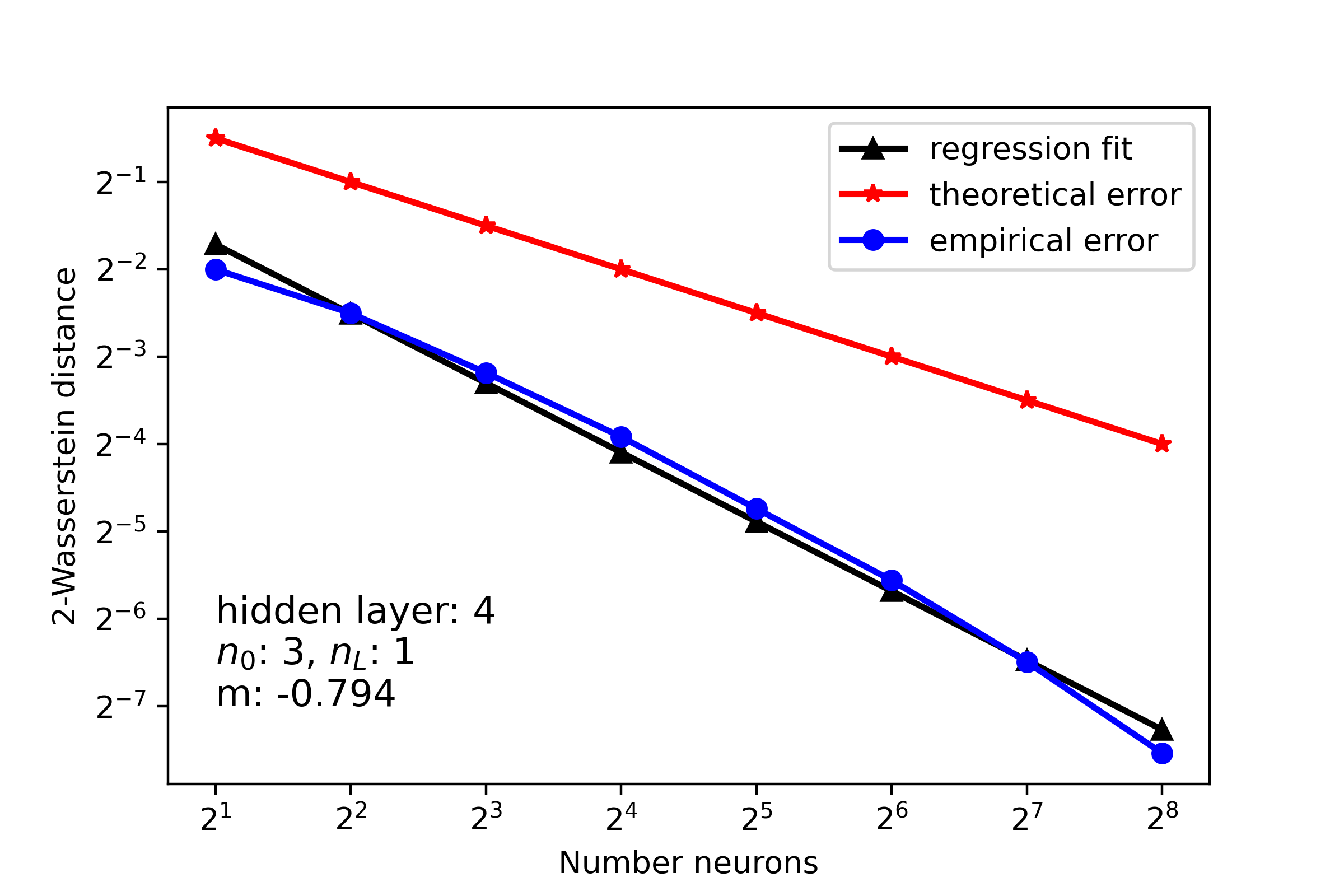}
    \caption{
    The above plots refer to the case of a  fixed single input, namely $[1,1,1]\in\R^3$, and a one-dimensional output $n_L=1$, while letting the number of hidden layers vary between $1$ and $4$. The fitted lines (black) even suggest a better rate of convergence. Image produced by the authors as described in Section \ref{sec:numerics}. }
    \label{fig:one-dim}
\end{figure}

\begin{figure}[h]
    \centering
    \includegraphics[width=.45\textwidth]{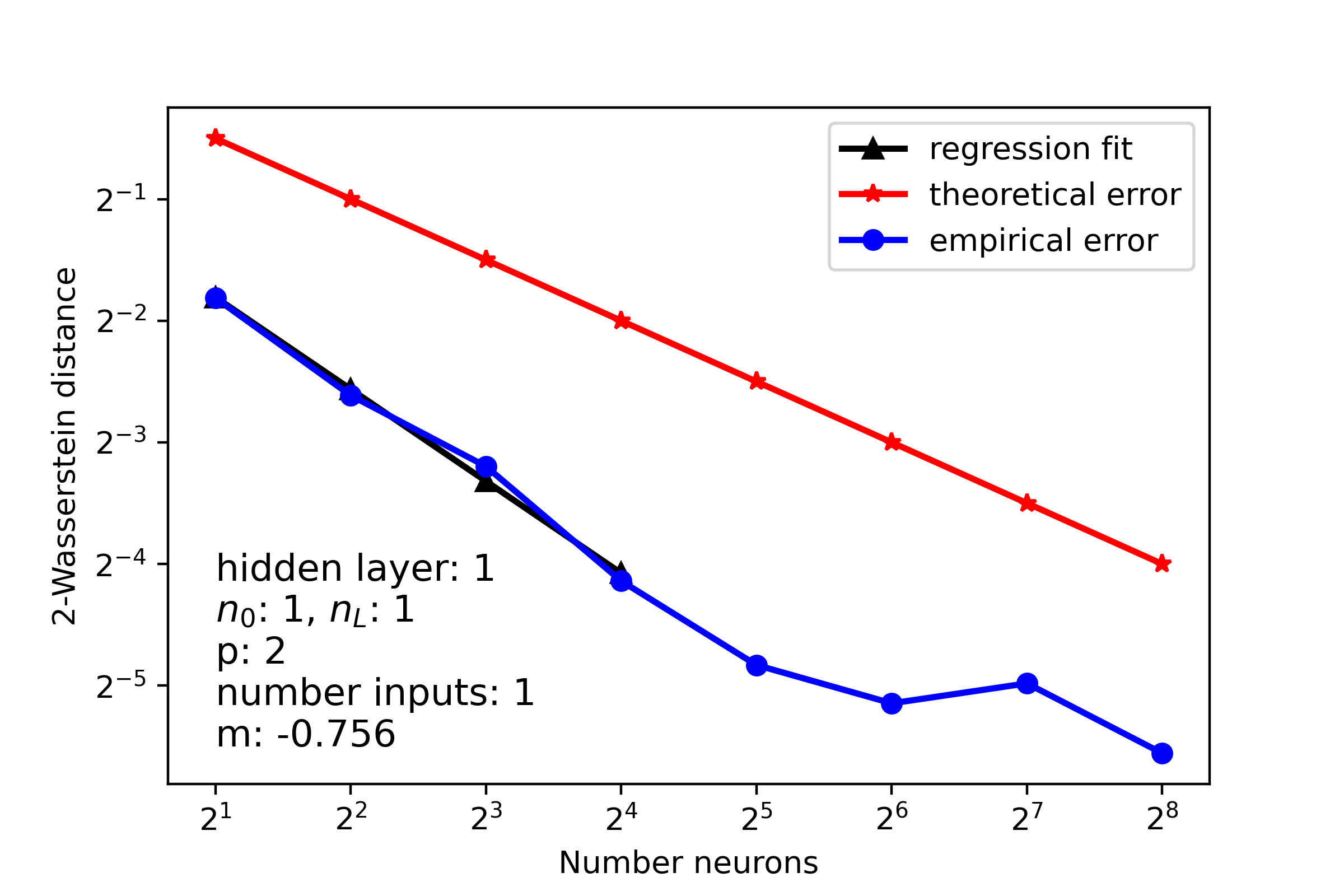}
    \includegraphics[width=.45\textwidth]{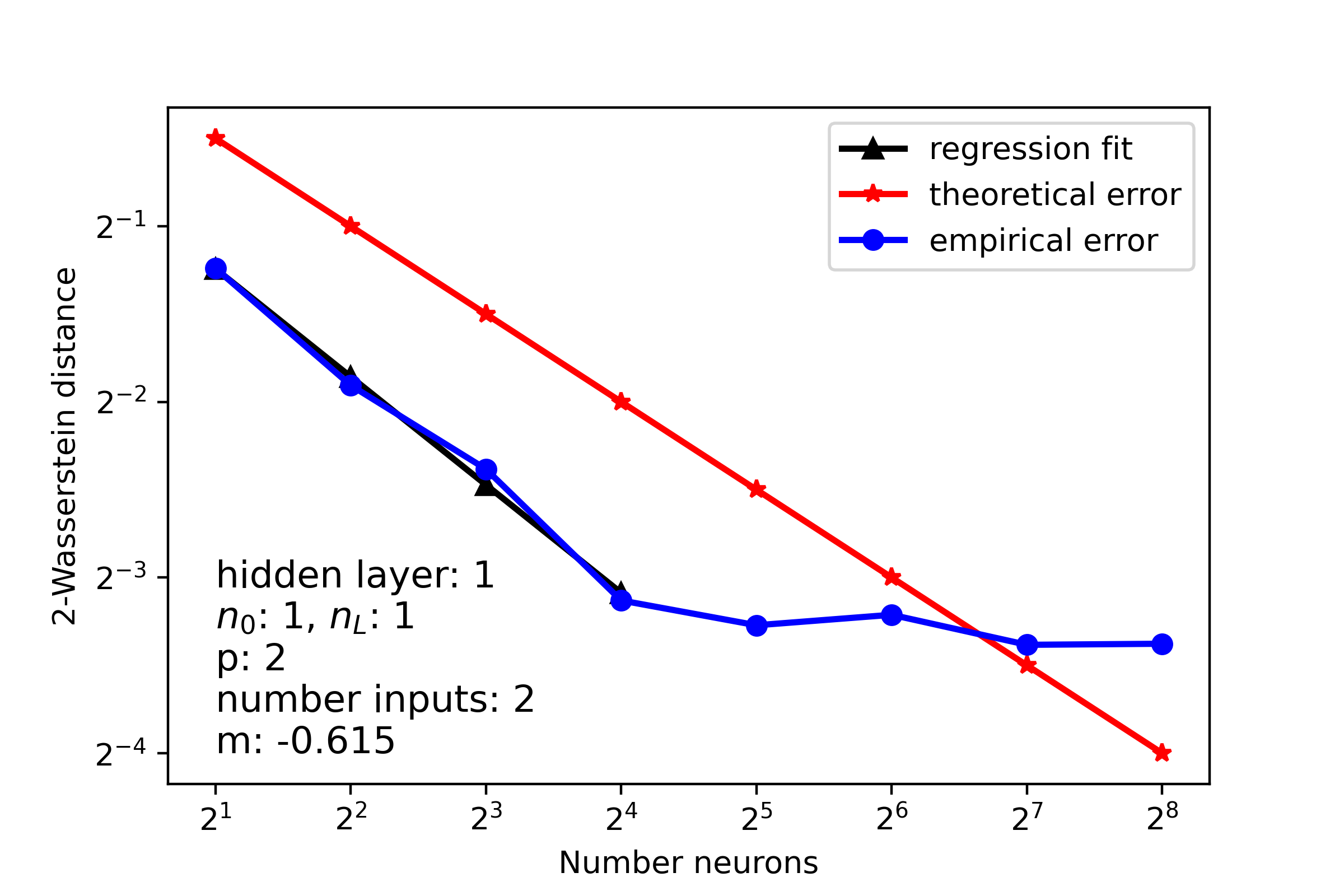}
    \includegraphics[width=.45\textwidth]{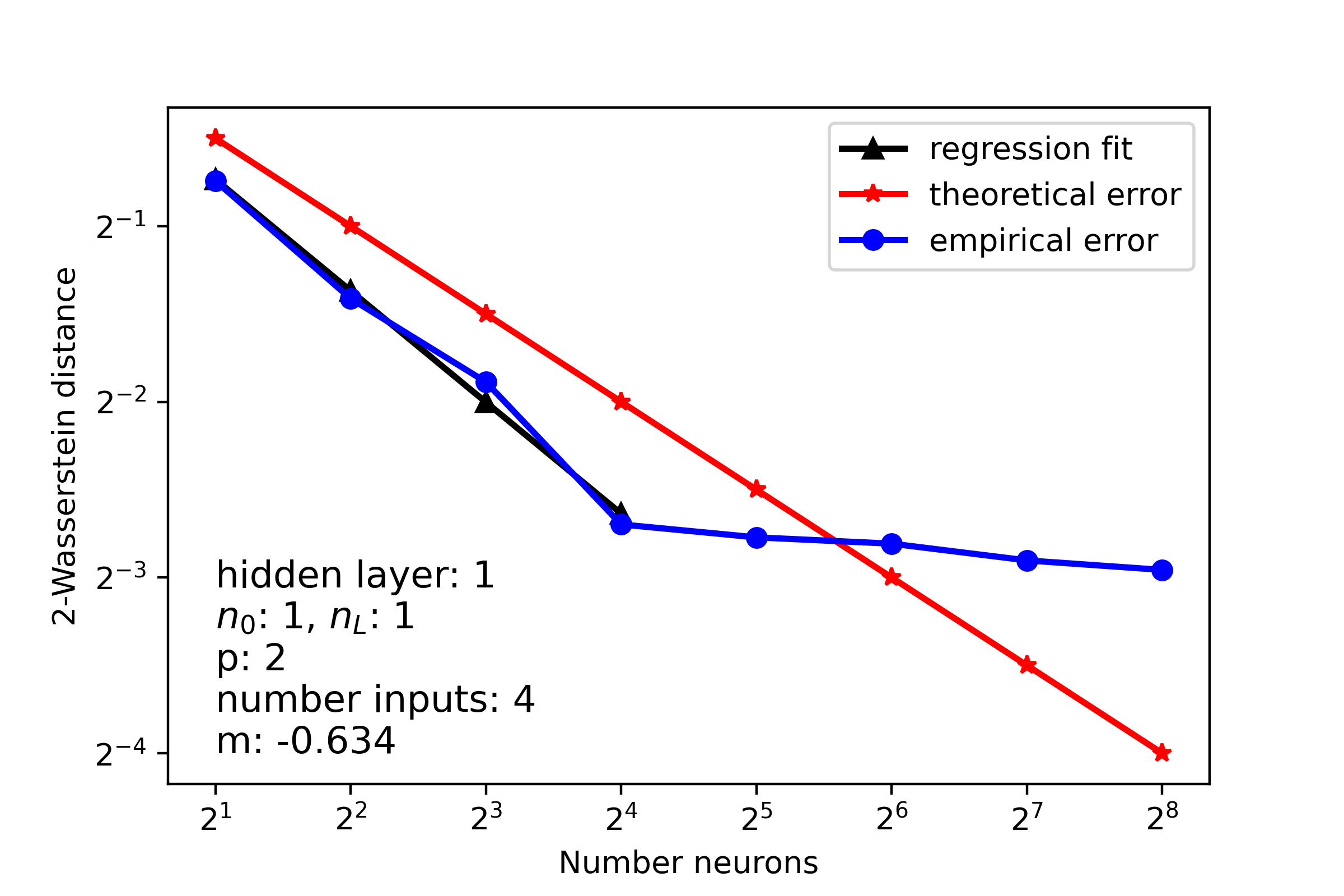}
    \includegraphics[width=.45\textwidth]{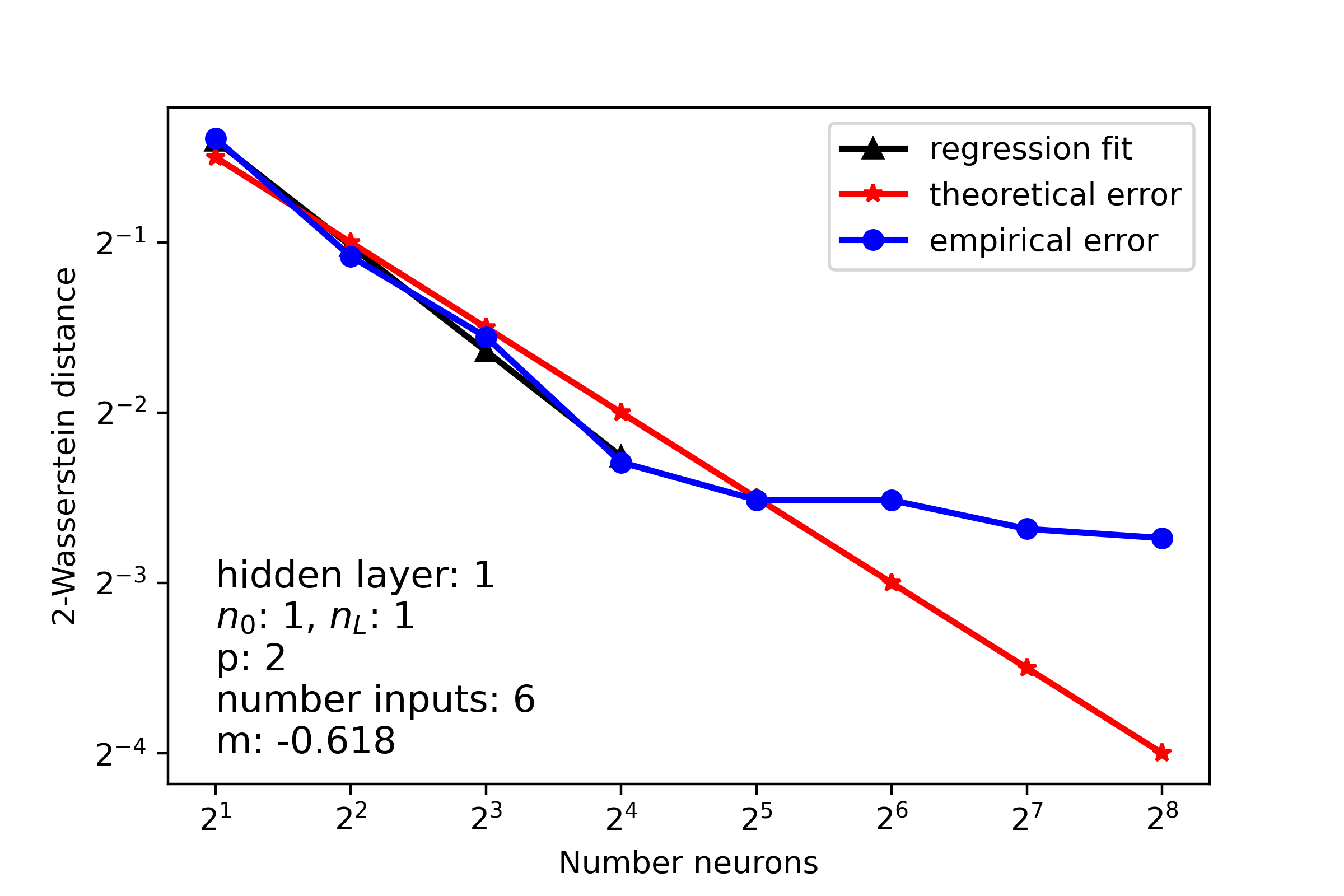}
    \caption{
    Keeping a single hidden layer and letting instead the number of inputs vary between $1$ and $6$ show that the blue lines flatten, because one would need a much larger sample to better approximate the empirical distance with the theoretical one. However, performing a regression on the first half of the graph still suggests convergence rates in agreement with our results. Image produced by the authors as described in Section \ref{sec:numerics}.
    }
    \label{fig:more-inputs}
\end{figure}

\begin{figure}[h]
    \centering
    \includegraphics[width=.45\textwidth]{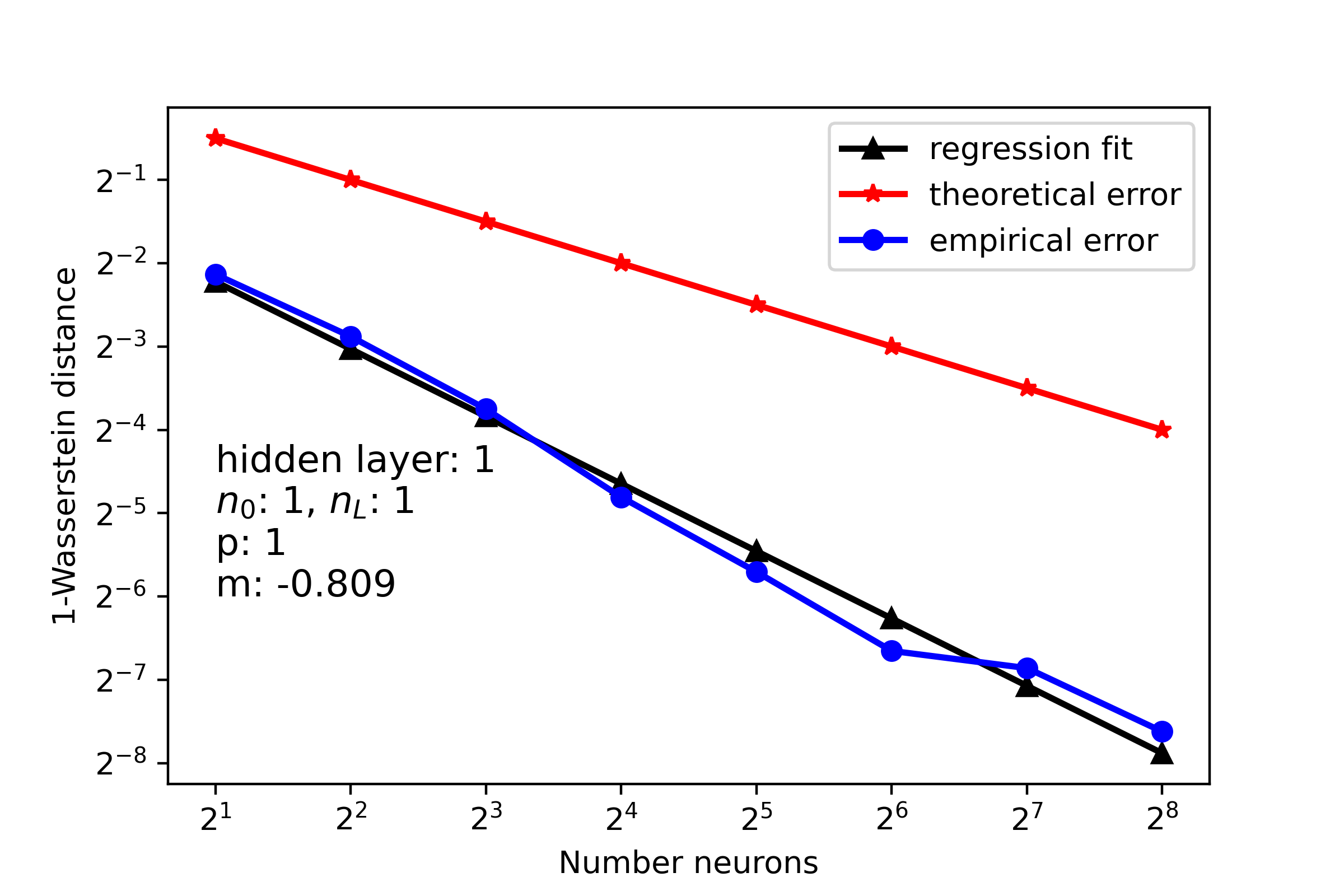}
    \includegraphics[width=.45\textwidth]{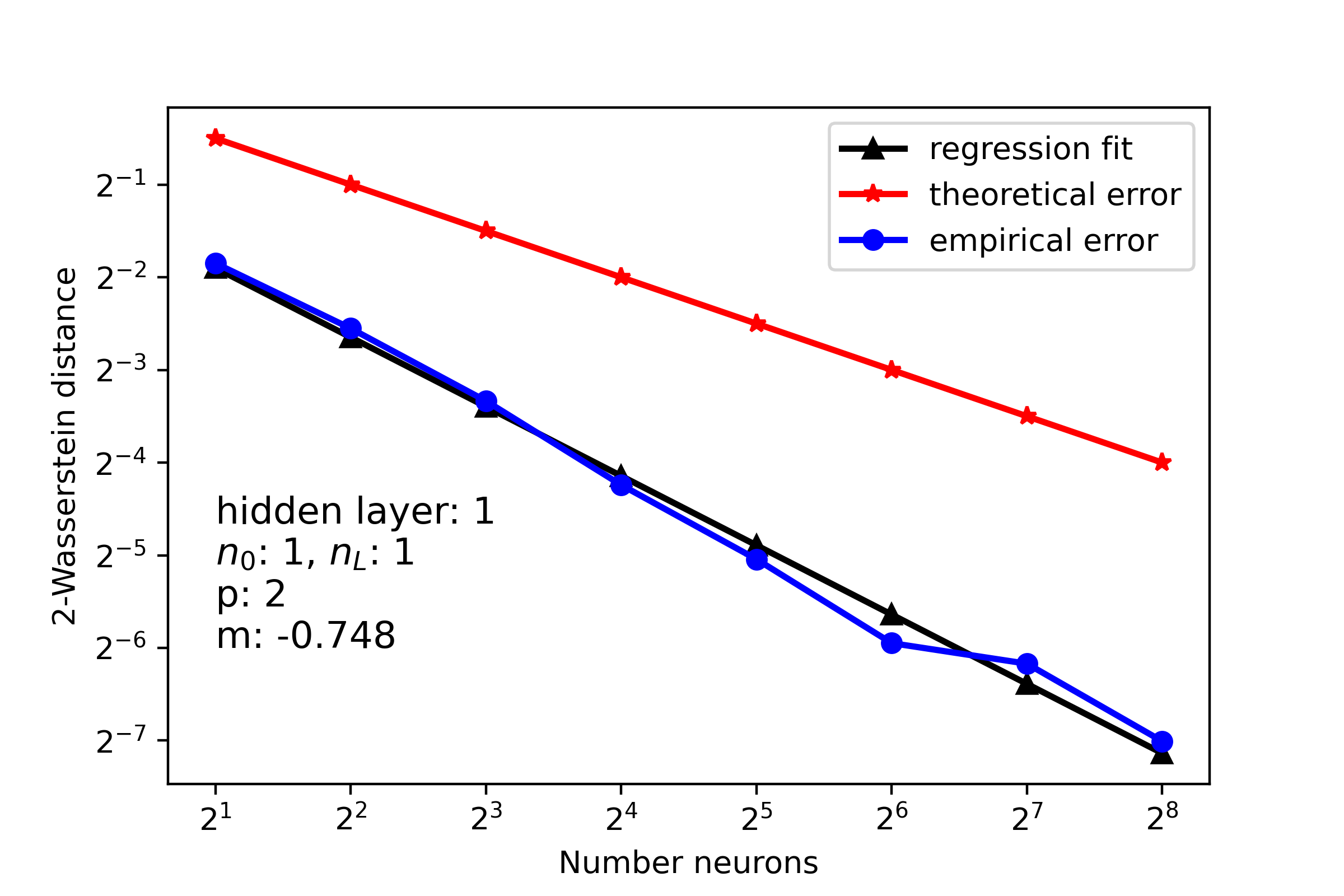}
    \includegraphics[width=.45\textwidth]{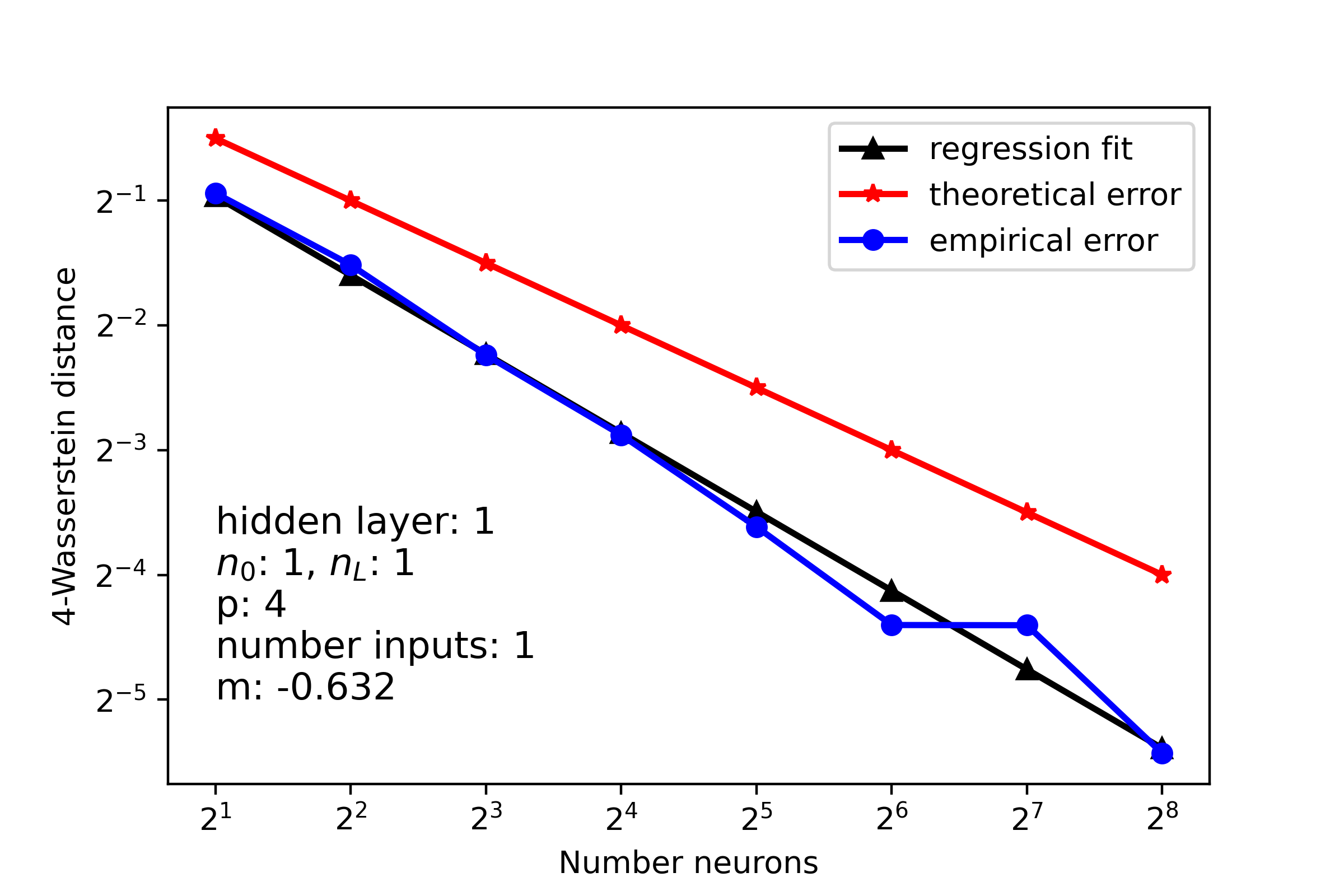}
    \includegraphics[width=.45\textwidth]{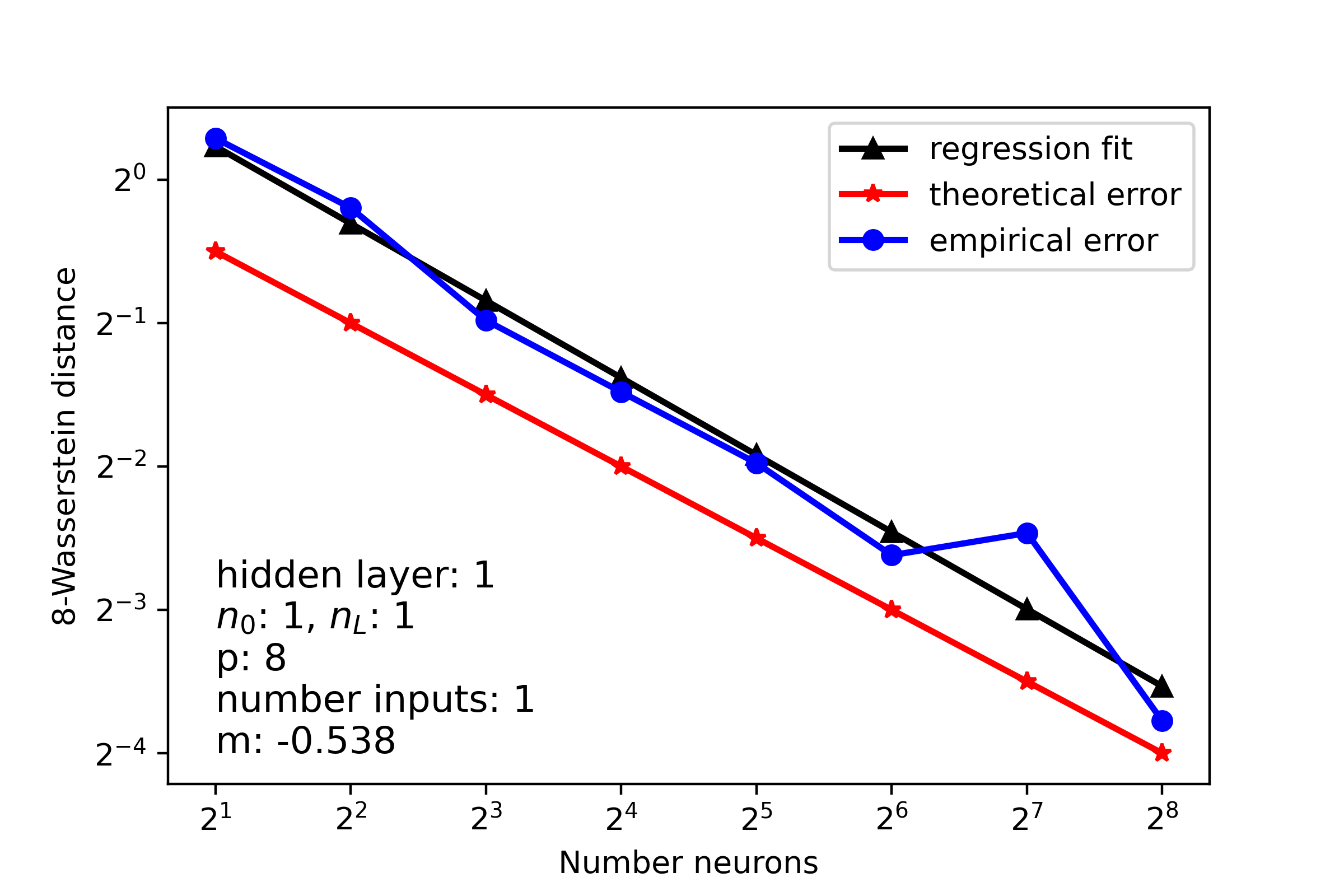}

    \caption{
    In the plots above, we fix a single input and single hidden layer structure and compute the Wasserstein distances of order $p$, for $p \in \cur{1,2,4,8}$ (defined by replacing the square in the definition of $\W_2$ with a $p$-th power and taking the $p$-th root). 
    The slopes of the regression lines seem to decrease as $p$ grows, getting closer to $-1/2$. The multiplicative constant seems to increase as the black line surpasses the reference red line. Image produced by the authors as described in Section \ref{sec:numerics}.}
    \label{fig:pvaries1inputs}
\end{figure}

In this section we present some numerical simulations to investigate the sharpness of Theorem~\ref{thm:main-clean}. The results are shown in Fig. \ref{fig:one-dim}, Fig. \ref{fig:more-inputs} and Fig. \ref{fig:pvaries1inputs}. Each experiment is structured as follows:
\begin{enumerate}
\item  we fix the main parameters for a network architecture, i.e.\ the number of layers $L$, the input dimension $n_0$, output dimension $n_L$ and a set of inputs $\X = \cur{x_i}_{i=1, \ldots, k} \subseteq \R^{n_0}$,
\item we then set the number of neurons (i.e., the width) in the hidden layers as a parameter $n = n_1= \ldots = n_{L-1}$  (for simplicity we let them be all equal),
\item we sample a large number $N$ of random Gaussian neural networks with weights sampled according to \eqref{eq:variances-clean} (for simplicity we let the biases be identically zero), and for each sample $j \in \cur{1, \ldots, N}$ we compute the output $y_j \in \R^{n_L \times k}$ of the network at the inputs $\X$, and use the results to define the empirical distribution
\[ \mu_N =  \frac 1 N \sum_{j=1}^N \delta_{y_j}.\]
\item we sample $N$ i.i.d.\ points $\cur{z_j}_{j=1}^N$ from the centered multivariate Gaussian $\mathcal{N}(K^{(L)}[\X])$, and define a corresponding empirical distribution
\[ \nu_N = \frac 1 N \sum_{j=1}^N \delta_{z_j},\]
\item we finally compute the Wasserstein distance between the two empirical distributions,
\begin{equation}\label{eq:Was-empirical} \W_2\bra{ \mu_N , \nu_N}. \end{equation}
\end{enumerate} 

As $N$ grows, the empirical quantity \eqref{eq:Was-empirical} converges to the corresponding theoretical one, i.e., the left hand side of \eqref{eq:thesis-clean}. Precisely, using e.g.\ the bounds from \cite{fournier2015rate}, we can approximate
\[
 \abs{ \W_2\bra{ f^{(L)}[\X], G^{(L)}[\X]} - \W_2\bra{ \mu_N,\nu_N }} \le C N^{-\alpha},
\]
where $\alpha>0$ depends only on the dimension $n_0 \times k$ and $C$ is a suitable constant (depending on the dimension but also some mild properties of the laws, e.g.\ their moments). When $N$ is sufficiently large, so that the error term above is much smaller than the right hand side of \eqref{eq:thesis-clean}, we expect an inequality of the type
$$
\W_2\bra{ \mu_N,\nu_N } \le \frac{C}{\sqrt{n}},
$$
for some constant $C>0$.

As a rule of thumb, we need $N^{-\alpha} \ll n^{-1/2}$. It always holds $\alpha \le \frac{1}{2}$, as dictated by the CLT, although for higher dimensions the curse of dimensionality appears, making $\alpha$ smaller as the dimension $n_L \times k$ grows. In addition to this, the exact computation of \eqref{eq:Was-empirical} becomes rather time-consuming, and we settled for $N=10^5$ in our experiments. We used the Python library \emph{POT: Python Optimal Transport} \cite{flamary2021pot} to compute the Wasserstein distance \eqref{eq:Was-empirical} and the \emph{neural-tangents} module \cite{neuraltangents2020} to compute the kernel $K^{(L)}[\X]$.

In the plots below, the blue lines connect the computed Wasserstein distances between the empirical distributions (in logarithmic scale as a function of the width $n$ of the hidden layers), the black line is a regression line whose slope $m$ is reported at the corner of each plot. The red line is the function $n^{-1/2}$ and it is purely provided for reference. They all show a good agreement with our theoretical result, and may suggest even better rates of convergence, due to a fitted slope $m$ much smaller than $-1/2$.

\section{Declarations}
\noindent \textbf{Funding} - No funds, grants, or other support were received. \\
\textbf{Conflicts of interest/Competing interests} - Not applicable.\\
\textbf{Ethics approval} - Not applicable. \\
\textbf{Consent to participate} - Not applicable. \\
\textbf{Consent for publication} - The authors of this manuscript consent to its publication. \\
\textbf{Availability of data and material} - Not applicable. \\
\textbf{Code availability} - The code for is available upon request. Its implementation is discussed in Section \ref{sec:numerics}. \\
\textbf{Authors' contributions} - Both authors contributed equally to this work. \\

\printbibliography

\end{document}